# Incentivizing Multi-Tenant Split Federated Learning for Foundation Models at the Network Edge

Songyuan Li, Jia Hu, Geyong Min, and Haojun Huang

***Abstract*—Foundation models (FMs) such as GPT-4 exhibit exceptional generative capabilities across diverse downstream tasks through fine-tuning. Split Federated Learning (SFL) facilitates privacy-preserving FM fine-tuning on resource-constrained local devices by offloading partial FM computations to edge servers, enabling device-edge synergistic fine-tuning. Practical edge networks often host multiple SFL tenants to support diversified downstream tasks. However, existing research primarily focuses on single-tenant SFL scenarios, and lacks tailored incentive mechanisms for multi-tenant settings, which are essential to effectively coordinate self-interested local devices for participation in various downstream tasks, ensuring that each SFL tenant's distinct FM fine-tuning requirements (e.g., FM types, performance targets, and fine-tuning deadlines) are met. To address this gap, we propose a novel Price-Incentive Mechanism (PRINCE) that guides multiple SFL tenants to offer strategic price incentives, which solicit high-quality device participation for efficient FM fine-tuning. Specifically, we first develop a bias-resilient global SFL model aggregation scheme to eliminate model biases caused by independent device participation. We then derive a rigorous SFL convergence bound to evaluate the contributions of heterogeneous devices to FM performance improvements, guiding the incentive strategies of SFL tenants. Furthermore, we model inter-tenant device competition as a congestion game for Stackelberg equilibrium (SE) analysis, deriving each SFL tenant's optimal incentive strategy. Extensive simulations involving four representative SFL tenant types (ViT, BERT, Whisper, and LLaMA) across diverse data modalities (text, images, and audio) demonstrate that PRINCE accelerates FM fine-tuning by up to 3.07x compared to state-of-the-art approaches, while consistently meeting fine-tuning performance targets.**

## I. INTRODUCTION

WE have been witnessing unprecedented breakthroughs in foundation models (FMs) such as BERT [1] and LLaMA [2], which demonstrate near-human cognition capabilities and generate human-like responses for various applications, including personal assistants [3] and autonomous driving [4]. FM needs extensive pre-training on massive datasets to acquire broad, general-purpose knowledge, which can then be transferred to downstream tasks aimed at achieving domain-specific application goals (e.g., image classification, and sentiment analysis). To specialize in these downstream tasks, pre-trained FMs require fine-tuning with domain-specific data (e.g., user reviews, photos or emails), which is often privacy-sensitive and necessitates strict on-device storage and processing. However, many local devices, such as smartphones and smartwatches, are resource-constrained to handle computation-intensive FM fine-tuning workloads (e.g., a typical FM requires 250,000 Petaflops for full-parameter fine-tuning [5]). To overcome these issues, edge computing provides low-latency computation power at the network edge (e.g., edge servers) close to local devices, enabling the offloading of partial intensive FM computations from local devices to capable edge servers. Specifically, Split Federated Learning (SFL) [6] leverages edge computing for efficient, privacy-preserving FM fine-tuning, which alleviates the FM computation burden on local devices through device-edge synergistic fine-tuning, while preserving the data privacy on local devices.

SFL enables a collaborative FM fine-tuning paradigm in resource-constrained edge networks by integrating Federated Learning (FL) [7] and Split Learning (SL) [8]. It combines FL's ability to parallelize FM fine-tuning across multiple local devices without exposing their private data, with SL's model splitting approach to address local devices' resource constraints. In SFL, a FM is partitioned at a designated cut layer into device-side and server-side submodels, deployed on local devices and an edge server, respectively, for device-edge synergistic FM fine-tuning. Device-side submodels handle FM fine-tuning workloads tailored to local computational capabilities, while the remaining FM workloads are offloaded to the powerful edge server, reducing computation costs on resource-constrained devices. During FM fine-tuning, only intermediate activations/gradients of the cut layer are exchanged between local devices and the edge server, ensuring that local private data remains confidential. Similar to FL, SFL performs global FM synchronization at the edge server by aggregating FM updates from both device-side and server-side submodels.

In edge networks, an SFL tenant employs local devices to participate in SFL for a downstream task. Existing SFL studies [9][10][11][12][13] primarily focus on single SFL tenant scenarios, assuming that a single SFL tenant has exclusive control over all local devices. However, with the growing proliferation of real-world intelligent applications, practical edge networks often host multiple SFL tenants to accommodate diversified downstream tasks. These SFL tenants shares local devices and engages a subset of them for separate downstream tasks. For instance, social media platforms like Twitter and Instagram compete to leverage smartphone statistics for respective data analytic tasks, such as personalized content recommendations and trending topic predictions [14].

*Corresponding author: Jia Hu and Geyong Min.*

Songyuan Li, Jia Hu, and Geyong Min are with the Department of Computer Science, Faculty of Environment, Science and Economy, University of Exeter, Exeter EX4 4PY, U.K. (e-mail: S.Y.Li@exeter.ac.uk, J.Hu@exeter.ac.uk, G.Min@exeter.ac.uk).

Haojun Huang is with the School of Electronic Information and Communications, Huazhong University of Science and Technology, Wuhan 430074, China (e-mail: hjhuang@hust.edu.cn).

Furthermore, local devices are inherently self-interested in SFL. Without sufficient incentives from SFL tenants, local devices may not be willing to engage in the SFL tenant's downstream task due to the heavy workloads associated with local FM fine-tuning. Moreover, multiple SFL tenants often have diverse FM fine-tuning requirements, in terms of FM types, performance targets, and fine-tuning deadlines.

Therefore, it is imperative to have a tailored incentive mechanism that guides multiple SFL tenants to offer strategic price incentives, which solicit high-quality device participation for device-edge synergistic fine-tuning, thereby satisfying their FM fine-tuning performance requirements. However, the presence of multiple SFL tenants significantly complicates the incentive mechanism design, necessitating a thorough exploration of the following aspects:

- *Independent Device Participation*: Each local device independently decides whether to participate in downstream tasks based on the price incentives offered by various SFL tenants. This selective participation can cause certain downstream tasks to be dominated by data from a small subset of devices with disproportionately high participation levels, resulting in FM model biases for SFL tenants.
- *Device Contribution Assessment*: Each SFL tenant aims to implement an effective incentive strategy that rewards local devices contributing more to FM performance improvements. However, before completing the FM fine-tuning, it is difficult to accurately assess how a given local device's SFL participation will impact the final FM learning performance.
- *Inter-tenant Device Competition*: Since local devices are shared among multiple SFL tenants, these SFL tenants inherently compete for robust devices with high-quality training data. Consequently, it is crucial to effectively manage inter-tenant competition and ensure balanced FM fine-tuning performance across all SFL tenants.

To address these challenges, we develop a bias-resilient global SFL model aggregation scheme that guarantees the convergence to a globally optimal and unbiased FM model with arbitrary independent device participation levels. Building on this, we derive a rigorous SFL convergence bound that predicts the contributions of heterogeneous local devices to FM performance improvements, without requiring actual FM fine-tuning. This SFL convergence bound analytically evaluates the impact of varying SFL participation levels of local devices with non-independent and identically distributed (non-i.i.d.) data on global FM performance, which is used to guide the design of the pricing strategies of SFL tenants. Furthermore, the strategic interaction between SFL tenants and local devices is modeled as a multi-leader multi-follower Stackelberg game. Specifically, SFL tenants (leaders) determine their optimal pricing strategies to stimulate high-quality device participation, while local devices (followers) respond by selecting their self-interested device participation levels in downstream tasks. To coordinate inter-tenant device competition, each SFL tenant is modeled as a player in a congestion game, engaging in organized contention for robust local devices. In a nutshell, this paper makes the following contributions:

- To the best of our knowledge, we are the first to explore effectively incentivizing multi-tenant SFL for FM fine-tuning in edge networks. Self-interested local devices are incentivized to participate in downstream tasks of multiple SFL tenants at desired participation levels. Empowered by our bias-resilient SFL design, SFL tenants eliminate the FM model biases caused by *independent device participation*. The pricing strategies of SFL tenants are guided by *device contribution assessments* derived from our SFL convergence bound.
- We develop a novel Prince-Incentive mechanism for multi-tenant SFL, named PRINCE, based on the multi-leader multi-follower Stackelberg game. The *inter-tenant device competition* is managed through congestion game modeling, thereby balancing each SFL tenant's FM fine-tuning requirements. A decentralized price-incentive algorithm for multi-tenant SFL is proposed, achieving finite-time convergence to a Stackelberg Equilibrium (SE) solution that determines each tenant's optimal incentive strategy.
- Our PRINCE mechanism is evaluated through extensive simulations with four representative types of SFL tenants, each managing distinct FM fine-tuning workloads, ViT, BERT, Whisper, and LLaMA, across various data modalities including text, images, and audio. Compared with several state-of-the-art incentive approaches, the experimental results demonstrate that PRINCE accelerates FM fine-tuning for each SFL tenant's downstream task by up to 3.07x, while consistently meeting the performance targets of FM fine-tuning. The code of PRINCE is available at: https://github.com/songyuanli/AERIA.

The rest of this paper is organized as follows. Section II introduces the related work. Section III presents the multi-tenant SFL system. Section IV depicts our incentive-driven device participation game. Section V develops a decentralized implementation of our PRINCE mechanism. Section VI discusses the experimental results. Section VII concludes this paper.

## II. Related Work

**Split Federated Learning (SFL):** It is considered a variant of classical Federated Learning (FL) [7], specifically designed to address the limitation of operating FL on resource-constrained devices by offloading partial machine learning workloads to capable servers via layer-wise model split. Thapa *et al.* [6] was the first to identify the weaknesses of FL and introduce the SplitFed concept, which partitions deep neural network (DNN) workloads between the local device and the server. Lin *et al.* [9] explored efficient model splitting strategies to accelerate the SFL process at the resource-constrained network edge. Wu *et al.* [10] addressed the communication efficiency challenge in SFL, achieving significant reductions in communication overhead without compromising DNN accuracy targets. Gao *et al.* [11] identified device heterogeneity in computation and communication as a factor that could slow down the SFL process, and designed a pipelined SFL framework to accelerate SFL on heterogeneous devices. Fu *et al.* [12] tackled the challenges of system and statistical heterogeneity in SFL by proposing a joint scheme for adaptive model splitting and quality-aware device selection on various DNN tasks. Ganguly *et al.* [13] developed a network-aware SFL scheme that dynamically tuned the contributions of various network elements to the SFL process, adapting to evolving network conditions such as device mobility.

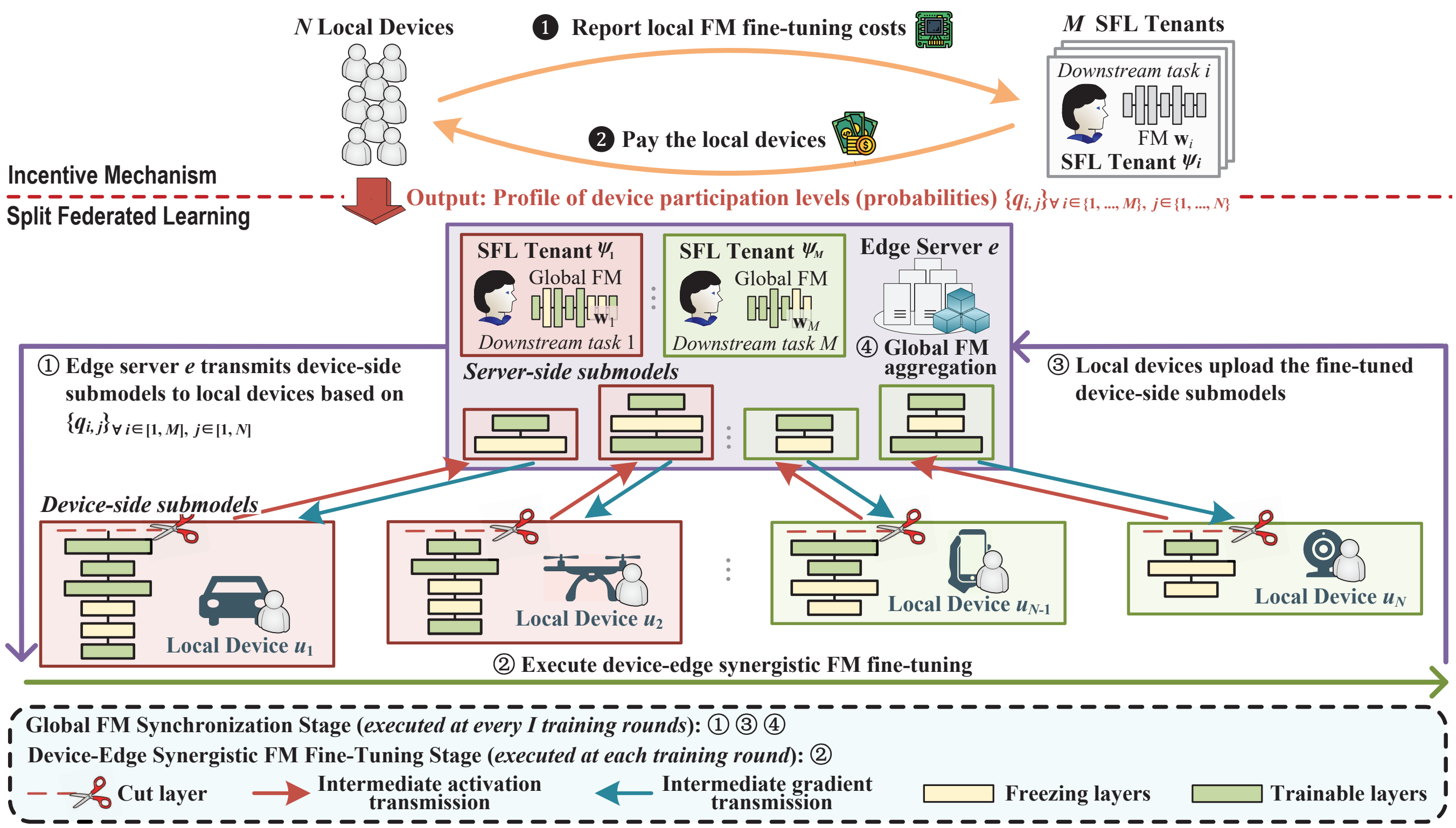

Fig. 1: Multi-tenant Split Federated Learning (SFL) system in edge networks, and its incentive mechanism design.

**Incentive Mechanism Design for Federated Learning:** It has emerged as a fundamental issue in FL, given the self-interested nature of local devices which require monetary incentives to encourage their FL participation. Thi Le *et al.* [15] implemented an auction-based incentive mechanism for FL, allowing devices to self-assess their local DNN training costs and bid for FL participation opportunities. Deng *et al.* [16] proposed a quality-aware incentive mechanism that accurately estimated the individual learning quality of local devices, enabling targeted incentives for robust devices. Luo *et al.* [17] designed an unbiased incentive mechanism that implemented optimal incentive strategies to motivate devices with varying FL participation levels, thereby achieving a globally optimal unbiased DNN model. Wang *et al.* [18] developed a dynamic pricing solution for device recruitment in FL, considering the recruitment process spanning a period of time and the need for adaptive adjustment of pricing compensation based on the actual arrival patterns of device clients. Liao *et al.* [19] proposed an incentive-compatible mechanism for heterogeneous client sampling, addressing key incentive challenges in FL, including information asymmetry and client strategic behaviors. Wang *et al.* [20] designed a price-based incentive mechanism to adaptively induce high-quality FL participation from local devices, thereby mitigating global DNN model bias and accelerating FL convergence.

Distinguishing from the existing SFL literature that is limited to DNN workloads, we are the first to extend SFL to FM fine-tuning workloads, and empirically demonstrate its efficacy across various downstream tasks with diverse data modalities (e.g., text, images, and audio). In light of the growing diversity of real-world intelligent applications, this research further fills a technical void in multi-tenant SFL by proposing an incentive mechanism to foster high-quality device participation in downstream tasks of multiple SFL tenants. The mechanism ensures that each SFL tenant's distinct FM fine-tuning requirements, in terms of FM types, performance targets, and fine-tuning deadlines, are satisfied.

## III. Multi-Tenant SFL System

### A. System Overview

As shown in Figure 1, we envision a multi-tenant SFL system over edge networks, accommodating $M$ SFL tenants $\Psi = \{\psi_i\}_{i=1}^M$ to adapt their FMs for diverse downstream tasks. The multi-tenant SFL framework fosters device participation in these downstream tasks through a novel incentive mechanism design, and accelerates FM fine-tuning via device-edge synergy. Specifically, our multi-tenant SFL framework comprises two key components:

**Incentive Mechanism for Device Participation:** Multiple SFL tenants $\psi_i$ coexist in edge networks, each managing its respective split federated FM fine-tuning workload for a distinct downstream task (indexed by $i$). These SFL tenants leverage monetary incentives to competitively recruit SFL participants from a shared pool of local devices $\mathcal{U} = \{u_j\}_{j=1}^N$, to contribute to their downstream tasks. To maximize FM performance on their respective downstream tasks, each SFL tenant $\psi_i$, with a payment budget $B_i$, develops a customized pricing strategy $\mathbf{P}_i = \{P_{i,1}, ..., P_{i,N}\}$ for local devices $u_i \in \mathcal{U}$, thereby incentivizing high-quality device participation. Each local device $u_j$ responds to the monetary incentive from SFL tenants by independently determining its device participation levels (probabilities) $\{q_{i,j}\}_{\forall i \in [1,M]}$ for various downstream tasks $i \in \{1, ..., M\}$, with the goal of maximize its device utility after deducting local FM fine-tuning costs. Local devices $u_j \in \mathcal{U}_i$ (where $\mathcal{U}_i \subseteq \mathcal{U}$) that participate in the SFL tenant $\psi_i$'s downstream task receive equitable monetary rewards based on their individual contributions to improving the learning performance of the FM $\mathbf{w}_i$.

**Split Federated Learning:** Local devices $u_j \in \mathcal{U}_i$ participating in the SFL tenant $\psi_i$'s downstream task allocate their local computation resources for fine-tuning the FM $\mathbf{w}_i$ in parallel, using their respective local training datasets $\mathcal{D}_{i,j}$. To mitigate the FM fine-tuning costs on resource-constrained devices, the FM fine-tuning process operates in each training round $r$ through device-edge synergy. By partitioning the FM $\mathbf{w}_i$ into device-side ($\mathbf{w}_{i,j}^C$) and server-side ($\mathbf{w}_{i,j}^S$) submodels, the local device $u_j$ handles partial FM fine-tuning workloads according to its computation capacity, while the primary FM fine-tuning workload is offloaded to the capable edge server $e$. Every $I$ training rounds, each SFL tenant $\psi_i$ performs global FM synchronization at the edge server $e$. This process involves aggregating model updates from the fine-tuned device-side and server-side submodels of all participating local devices $u_j \in \mathcal{U}_i$. Simultaneously, each local device $u_j \in \mathcal{U}$ chooses the downstream task to join for the next $I$ training rounds, based on its device participation levels $\{q_{i,j}\}_{\forall i \in [1,M]}$. The globally synchronized model parameters in $\mathbf{w}_i$ are then broadcast back to the participating devices $u_j \in \mathcal{U}_i$, which continue device-edge synergistic FM fine-tuning in the subsequent $I$ training rounds. The cycle of *device-edge synergistic FM fine-tuning* and *global FM synchronization* repeats until the global loss function converges or the FM fine-tuning deadline $\Gamma_i$ is reached.

*B. Split Federated Learning Process*

The SFL process in edge networks comprises two primary stages: 1) device-edge synergistic FM fine-tuning, and 2) global FM synchronization. The device-edge synergistic FM fine-tuning is executed in each training round $r$, while global model synchronization occurs every $I$ rounds.

**Device-Edge Synergistic FM Fine-Tuning Stage:** For the local device $u_j$ participating in the SFL tenant $\psi_i$'s downstream task, the FM $\mathbf{w}_i$ is partitioned at the cut layer $s_{i,j}$ into device-side ($\mathbf{w}_{i,j}^C$) and server-side ($\mathbf{w}_{i,j}^S$) submodels, i.e.,

$$\mathbf{w}_{i,j}^C = \left\{\mathbf{w}_{i,j}^n\right\}_{n=1}^{s_{i,j}}, \mathbf{w}_{i,j}^S = \left\{\mathbf{w}_{i,j}^n\right\}_{n=s_{i,j}+1}^{H_i}. \tag{1}$$

The device-edge synergistic FM fine-tuning stage, involving device-side and server-side submodel fine-tuning in each training round $r$, comprises the following five steps:

$\diamond$ *Device-side submodel forward propagation (Step 1):* All participating devices $u_j$ execute forward propagation on their respective device-side submodels $\mathbf{w}_{i,j}^C$ in parallel. Specifically, each local device $u_j$ leverages its local dataset $\mathcal{D}_{i,j}$ to fine-tune the device-side submodel $\mathbf{w}_{i,j}^C$. The local dataset $\mathcal{D}_{i,j} = \{x_{i,j}^d, y_{i,j}^d\}_{d=1}^{D_{i,j}}$ contains $D_{i,j}$ domain-specific data samples tailored for the SFL tenant $\psi_i$'s downstream task, where $x_{i,j}^n$ represents the $n$-th input data and $y_{i,j}^n$ denotes its corresponding ground-truth label. After feeding these data samples into $\mathbf{w}_{i,j}^C$ via forward propagation, intermediate activations $\mathbf{A}_{i,j}$ are generated at the cut layer $s_{i,j}$:

$$\mathbf{A}_{i,j} = \varphi\left(\mathbf{X}_{i,j}; \mathbf{w}_{i,j}^C\right) \tag{2}$$

where $\mathbf{X}_{i,j} = \{x_{i,j}^k\}_{k=1}^{D_{i,j}}$, and $\varphi(x; w)$ represents the mapping function which relates the data input $x$ to its predicted value given the model parameter $w$.

$\diamond$ *Intermediate activation transmission (Step 2):* Once the device-side submodel's forward propagation is completed, each local device $u_j$ uploads its generated activations $\mathbf{A}_{i,j}$ along with the ground-truth labels $\mathbf{Y}_{i,j} = \{y_{i,j}^n\}_{n=1}^{D_{i,j}}$ to the edge server $e$ via wireless networks. The edge server $e$ then utilizes these collected intermediate activations to fuel server-side submodel fine-tuning.

$\diamond$ *Server-side submodel forward and backward propagation (Step 3):* The edge server $e$ fine-tunes the server-side submodel $\mathbf{w}_{i,j}^S$ upon receiving the intermediate activations $\boldsymbol{A}_{i,j}$ from the local device $u_j$. By feeding the collected activations $\mathbf{A}_{i,j}$ into $\mathbf{w}_{i,j}^S$, the edge server $e$ executes the server-side forward propagation and computes the predicted labels $\widetilde{y}_{i,j}$:

$$\hat{\mathbf{y}}_{i,j} = \varphi\left(\mathbf{A}_{i,j}; \mathbf{w}_{i,j}^S\right) \tag{3}$$

The predicted labels $\widetilde{\mathbf{y}}_{i,j}$ are used to calculate loss function value against the ground-truth labels $\mathbf{y}_{i,j}$, which further derives the server-side submodel's loss gradients through backward propagation.

Let $\widetilde{\mathbf{w}}_{i,j}^S$ denote the set of server-side trainable layers, and we define $\nabla F_j(\widetilde{\mathbf{w}}_{i,j}^S; \mathbf{A}_{i,j})$ as the loss gradient at these server-side trainable layers, given the activation input $\mathbf{A}_{i,j}$. Therefore, the server-side submodel $\mathbf{w}_{i,j}^S$ can be updated through:

$$\widetilde{\mathbf{w}}_{i,j}^S \leftarrow \widetilde{\mathbf{w}}_{i,j}^S - \gamma \cdot \nabla F_j(\widetilde{\mathbf{w}}_{i,j}^S; \mathbf{A}_{i,j}), \tag{4}$$

where $\gamma$ is the learning rate. As the output of backward propagation on the server-side submodel $\mathbf{w}_{i,j}^S$, the intermediate gradients $\nabla F_j(\mathbf{w}_{i,j}^{s_{i,j}+1}; \mathbf{A}_{i,j})$ are generated at the cut layer.

$\diamond$ *Intermediate gradient transmission (Step 4):* Once the server-side submodel's backward propagation is completed, the edge server $e$ sends the intermediate gradients $\nabla F_i(\mathbf{w}_{i,j}^{s_{i,j}+1}; \mathbf{A}_{i,j})$ back to the corresponding local device $u_i$ for updating the device-side submodel $\boldsymbol{w}_{i,j}^C$.

$\diamond$ *Device-side submodel backward propagation (Step 5):* Upon receiving the intermediate gradients $\nabla F_j(\mathbf{w}_{i,j}^{s_{i,j}+1}; \mathbf{A}_{i,j})$, each participating local device $u_i$ updates its device-side submodel $\mathbf{w}_{i,j}^C$ through executing model backward propagation. Let $\widetilde{\mathbf{w}}_{i,j}^C$ represent the set of device-side trainable layers, and the device-side submodel $\mathbf{w}_{i,j}^C$ can be updated through:

$$\widetilde{\mathbf{w}}_{i,j}^C \leftarrow \widetilde{\mathbf{w}}_{i,j}^C - \gamma \cdot \nabla F_j(\widetilde{\mathbf{w}}_{i,j}^C; \mathbf{X}_{i,j}), \tag{5}$$

where $\nabla F_j(\widetilde{\mathbf{w}}_{i,j}^C; \mathbf{X}_{i,j})$ indicates the loss gradient at the device-side trainable layers $\widetilde{\mathbf{w}}_{i,j}^C$ given the data input $\mathbf{X}_{i,j}$.

**Global FM Synchronization Stage:** The edge server $e$ periodically executes global FM synchronization, separating SFL synchronization cycles. Every $I$ consecutive training rounds, each SFL tenant $\psi_i$ derives a new global FM $\mathbf{w}_j$ by aggregating the latest FM fine-tuning updates. Each local device $u_j \in \mathcal{U}_i$ participating in the SFL tenant $\psi_i$'s downstream task uploads its latest fine-tuned device-side submodels $\mathbf{w}_{i,j}^C$ to the edge server $e$. The edge server $e$ then pairs and assembles these collected device-side submodels $\mathbf{w}_{i,j}^C$ with the corresponding server-side submodels $\mathbf{w}_{i,j}^S$ to forge the device models $\mathbf{w}_{i,j}$ of each participating local device $u_i$:

$$\mathbf{w}_{i,j} \leftarrow \left(\mathbf{w}_{i,j}^C \parallel \mathbf{w}_{i,j}^S\right), \forall u_j \in \mathcal{U}_i. \tag{6}$$

TABLE I: Key Notations and Definitions

| Symbol | Definition |
|---|---|
| $\Psi, M, \psi_i$ | Set of SFL tenants, total number of SFL tenants, and the $i$-th SFL tenant. |
| $\mathcal{U}, N, u_j$ | Set of local devices, total number of local devices, and the $j$-th local device. |
| $I$ | Number of training rounds in each SFL synchronization cycle. |
| $\mathbf{w}_i$ | Global FM model of the SFL tenant $\psi_i$. |
| $\rho_{i,n}, \nu_{i,n}$ | Computation overhead for forward and backward propagation of the $n$-th layer in $\mathbf{w}_i$. |
| $h_{i,n}, g_{i,n}$ | Size of intermediate activations and intermediate gradients at the $n$-th layer in $\mathbf{w}_i$. |
| $\xi_i^S$ | Computation capacity provisioned at the edge server $e$ for the SFL tenant $\psi_i$'s downstream task. |
| $\xi_j^C$ | Computation capacity of local device $u_j$. |
| $D_{i,j}$ | Number of local training data samples at the local device $u_j$ for fine-tuning the FM $\mathbf{w}_i$. |
| $B_i$ | Payment budget of SFL tenant $\psi_i$ to incentivize SFL participation from local devices. |
| $\mathbf{P}_i = \{P_{i,1}, ..., P_{i,N}\}$ | Pricing strategy of SFL tenant $\psi_i$ customized for each local device $u_j \in \mathcal{U}$. |
| $\mathbf{q}_i = \{q_{i,1}, ..., q_{i,N}\}$ | Device participation levels (probabilities) of the SFL tenant $\psi_i$. |
| $\mathcal{U}_i$ | Set of local devices that participate in the SFL tenant $\psi_i$'s downstream task, where $\mathcal{U}_i \subseteq \mathcal{U}$. |
| $\mathbf{w}_{i,j}^C, \mathbf{w}_{i,j}^S$ | Device-side, and server-side submodels of local device $u_j$ participating in the SFL tenant $\psi_i$'s downstream task. |
| $\Gamma_i$ | FM fine-tuning deadline of the SFL tenant $\psi_i$. |
| $K_i^\Gamma$ | Number of SFL synchronization cycles undergone within the FM fine-tuning deadline $\Gamma_i$. |
| $\mathbf{w}_i^\Gamma$ | Global FM model obtained by the SFL tenant $\psi_i$ until the FM fine-tuning deadline $\Gamma_i$. |

In a manner akin to to the de facto FedAvg algorithm, these forged device models $\mathbf{w}_{i,j}$ of all $u_j \in \mathcal{U}_i$, which serves the SFL tenant $\psi_i$, are aggregated to obtain the new global FM $\mathbf{w}_i$. The split federated FM fine-tuning process of SFL tenant $\psi_i$ will operate on this new global foundation model $\mathbf{w}_i$ for the next $I$ training rounds.

**SFL Process Completion:** For each SFL tenant $\psi_i$, the loop of *device-edge synergistic FM fine-tuning* and *global model synchronization* continues until the global loss function $F(\mathbf{w}_j)$ converges, within the FM fine-tuning deadline $\Gamma_j$. The global loss function $F(\mathbf{w}_j)$ integrates the local losses of all participating devices $u_j \in \mathcal{U}_i$ over their respective local datasets $\mathcal{D}_{i,j}$, as follows:

$$F(\mathbf{w}_i) = \sum_{u_j \in \mathcal{U}_i} \left(a_{i,j} \cdot F_j(\mathbf{w}_i; \mathbf{X}_{i,j})\right) \tag{7}$$

where $a_{i,j} = D_{i,j} / \sum_{u_{j'} \in \mathcal{U}_i} D_{i,j'}$ denotes the weight of local device $u_j$ amongst $\mathcal{U}_i$, ensuring $\sum_{u_j \in \mathcal{U}_i} a_{i,j} = 1$. The convergence of global loss function $F(\mathbf{w}_j)$ indicates that the FM $\mathbf{w}_i$ has achieved the optimal learning performance, resulting in the optimal fine-tuned FM $\mathbf{w}_i^*$, which satisfies:

$$\mathbf{w}_i^* = \arg\min_{\mathbf{w}_i} F(\mathbf{w}_i). \tag{8}$$

**SFL Time Cost Analysis:** Let $\rho_{i,n}$ and $\nu_{i,n}$ denote the computation overhead for forward and backward propagation, respectively, of the $n$-th layer in the FM $\mathbf{w}_i$. Meanwhile, $h_{i,n}$ and $g_{i,n}$ represent the size of intermediate activations and intermediate gradients, respectively, for the $n$-th layer in $\mathbf{w}_i$. Each local device $u_j$ has a local computation capacity of $\xi_j^C$, and communicates with the edge server $e$ at wireless data upload and download rates of $\zeta_j^{\mathrm{u}}$ and $\zeta_j^{\mathrm{d}}$, respectively. For the downstream task of SFL tenant $\psi_i$, the provisioned edge computing resources have a capacity of $\xi_i^S$.

$\diamond$ *Communication time cost:* Wireless communication between local devices and the edge server $e$ encompasses the transmission of intermediate activations and gradients, as well as the global synchronization of device-side submodels. Therefore, the communication time cost between the local device $u_j$ and the edge server $e$ within one SFL synchronization cycle is formulated as:

$$T_{i,j}^{com} = \overbrace{\frac{I \cdot D_{i,j} \cdot h_{i,s_{i,j}} + |\mathbf{w}_{i,j}^C|}{\zeta_j^{\mathrm{u}}}}^{\text{Wireless Upstream}} + \overbrace{\frac{I \cdot D_{i,j} \cdot g_{i,s_{i,j}} + |\mathbf{w}_{i,j}^C|}{\zeta_j^{\mathrm{d}}}}^{\text{Wireless Downstream}}, \tag{9}$$

where $|\mathbf{w}_{i,j}^C|$ is the size of device-side submodel.

$\diamond$ *Computation time cost:* In each training round, the local device $u_j$ conducts forward propagation on its device-side submodel $\mathbf{w}_{i,j}^C$ until the cut layer $s_{i,j}$, then executes the backward propagation to adjust the model parameters in $\mathbf{w}_i^C$. The computation time cost for fine-tuning the device-side submodel in one training round is:

$$T_{i,j}^C = D_{i,j} \cdot \sum\nolimits_{n=1}^{s_{i,j}} \left(\rho_{i,n} + \nu_{i,n}\right) / \xi_j^C. \tag{10}$$

Meanwhile, the edge server $e$ conducts forward and backward propagation to update the server-side submodel $\mathbf{w}_{i,j}^S$ based on the intermediate activations received from the local device $u_j$. Correspondingly, the computation time cost for fine-tuning the server-side submodel in one training round is:

$$T_{i,j}^S = D_{i,j} \cdot \sum\nolimits_{n=s_{i,j}+1}^{H_i} \left(\rho_{i,n} + \nu_{i,n}\right) / \xi_i^S. \tag{11}$$

Finally, the total time cost for all local devices $u_j \in \mathcal{U}_i$ to collaborate on the SFL tenant $\psi_i$'s downstream task within one SFL synchronization cycle is derived as:

$$T_i = T^{\text{agg}} + \max_{u_j \in \mathcal{U}_i} \left( I \cdot \left( T_{i,j}^{C} + T_{i,j}^{S} \right) + T_{i,j}^{com} \right), \quad (12)$$

where $T^{\text{agg}}$ represents the time cost of global model aggregation. The overall time cost is determined by the maximum time taken by any local device $u_j \in \mathcal{U}_i$. Accordingly, the FM fine-tuning deadline $\Gamma_i$ can be represented in terms of the number of undergone SFL synchronization cycles as $K_i^{\Gamma} = \Gamma_i / T_i$.

### C. Incentive Mechanism Design for Device Participation

**Overall Mechanism Design:** As the local devices work as independent decision-makers who base their SFL participation on their own interests, we will investigate the impact of their independent decision-making characteristics on our incentive mechanism design. Specifically, multiple SFL tenants $\psi_i$ first estimate the potential contributions of local devices to their respective downstream task. Local devices with robust computation capacity, allowing them to handle more device-side FM fine-tuning workloads, or those with high-quality training datasets, would be more competent to accelerate the model performance. Based on the varying potential contributions of these local devices, each SFL tenant $\psi_i$ then designs an appropriate pricing strategy $\mathbf{P}_i = \{P_{i,1}, ..., P_{i,N}\}$ for each local device $u_j \in \mathcal{U}$ to incentivize their SFL participation. In the following, we will formulate the sequential decision problems between the SFL tenants and the local devices.

**SFL Tenant's Decision Problem:** Each SFL tenant $\psi_i \in \Psi$ strives to minimize its global FM loss $F(\mathbf{w}_i)$, as formulated in Eq. (8), within the FM fine-tuning deadline $\Gamma_i$. To achieve this, the SFL tenant $\psi_i$ imposes a pricing strategy $\mathbf{P}_i = \{P_{i,1}, ..., P_{i,N}\}$ within a payment budget $B_i$ to incentivize the desired device participation levels $q_{i,j}$.

Let $\mathbf{w}_i^{\Gamma}(\mathbf{q}_i)$ denote the global FM obtained by SFL tenant $\psi_i$ until the FM fine-tuning deadline $\Gamma_i$, when local devices $u_j \in \mathcal{U}$ participate in the downstream task $i$ with levels $\mathbf{q}_i$ under pricing strategy $\mathbf{P}_i$. Therefore, the SFL tenant $\psi_i$'s decision problem can formulated as the following **P1**:

$$(\mathbf{P1}): \min_{\mathbf{P}_i} \Lambda_i = \mathbb{E}\left[ F\left( \mathbf{w}_i^{\Gamma}(\mathbf{q}_i) \right) \right] \quad (13)$$

$$s.t. \quad \sum\nolimits_{j=1}^{N} P_{i,j} \leq B_i \quad (13a)$$

The objective function in Eq. (13) aims to minimize the expected global loss $\Lambda_i = \mathbb{E}[F(\mathbf{w}_i^{\Gamma}(\mathbf{q}_i))]$ for SFL tenant $\psi_i$. The randomness of global FM loss arises from probabilistic device participation $\mathbf{q}_i$ and the stochastic gradient descent-based FM fine-tuning process. The payment budget constraint in Eq. (13a) necessitates careful design of the optimal pricing strategy $\mathbf{P}_i$.

**Local Device's Decision Problem:** Each local device $u_j \in \mathcal{U}$ seeks to maximize its own utility $\lambda_j$ by choosing its optimal device participation levels $\{q_{i,j}\}_{i=1}^{M}$ for various downstream tasks. The device utility $\lambda_j$ is based on the received payment $P_{i,j} \cdot q_{i,j}$, after deducting the local training cost $C_{i,j}$ for its SFL participation in downstream tasks $i \in \{1, ..., M\}$.

The local training cost $C_{i,j}$ involves the resource consumption expenses incurred by device-side submodel computation. Intuitively, $C_{i,j}$ is positively correlated with the amount of device-side resource usage, e.g., it scales linearly with the computation overhead required for model training [12][14]. Meanwhile, a higher device participation level $q_{i,j}$ typically results in a higher local FM fine-tuning cost for the SFL tenant $\psi_i$'s downstream task. Therefore, we define the local FM fine-tuning cost function $C_{i,j}$ as:

$$C_{i,j} = c_i \cdot q_{i,j}^{\tau}, \ \tau \geq 1 \quad (14)$$

where the parameter $c_i$ is a cost coefficient that typically arises from the energy consumption of local devices [21][22]. As is standard in economic cost modelling [23][24], the exponent $\tau \geq 1$ represents a board class of convex cost functions, implying an increasing rate of cost as $q_{i,j}$ rises. The concrete form of cost functions $C_{i,j}$ cannot affect the effectiveness of our incentive mechanism design. Accordingly, we formulate each local device $u_j$'s decision problem as the following **P2**:

$$(\mathbf{P2}): \max_{\{q_{i,j}\}_{\forall i \in [1,M]}} \lambda_j = \sum_{\psi_i \in \Psi} (q_{i,j} \cdot P_{i,j}) - \sum_{\psi_i \in \Psi} C_{i,j} \quad (15)$$

$$s.t. \quad 0 \leq \sum_{\psi_i \in \Psi} q_{i,j} \leq 1 \quad (15a)$$

$$0 \leq q_{i,j} \leq 1, \ \forall \psi_i \in \Psi \quad (15b)$$

*Remark:* The SFL tenants and the local devices pursue divergent interest objectives, as detailed in the decision problems **P1** and **P2**. Multiple SFL tenants $\psi_i \in \Psi$ expect to minimize the global FM loss for their respective downstream task $i$, by imposing a pricing strategy $\mathbf{P}_i$ that offers optimal incentives to encourage device participation. Accordingly, each local device $u_j \in \mathcal{U}$ responds to the price incentives by choosing its optimal device participation levels for various downstream tasks, thus maximizing its own device utility $\lambda_j$. In this sequential decision-making process, finding the joint optimal solution for pricing strategy and device participation levels to align the divergent interest objectives of SFL tenants and local devices inherently suffers from significant computational intractability.

## IV. Incentive-driven Device Participation Game

### A. Stackelberg Game Formulation

To address the computational intractability in our incentive mechanism, we propose *a multi-leader multi-follower Stackelberg game* that coordinates the divergent interest objectives of SFL tenants and local devices. Stackelberg game theory offers a natural framework for describing sequential decision problems between different parties, enabling a deep analytical understanding of multi-agent optimization [14][25]. As illustrated in Figure 2, the sequential decision-making process between the SFL tenants and the local devices is structured into these two phases:

- **Game Phase I:** The SFL tenants $\psi_i \in \Psi$ act as the Stackelberg leades who determine their respective pricing strategies $\mathbf{P}_i = \{P_{i,1}, ..., P_{i,N}\}$ to minimize the expected global FM loss $\Lambda_i = \mathbb{E}\big[F\big(\mathbf{w}_i^{\Gamma}(\mathbf{q}_i)\big)\big]$, as defined in Problem **P1**.
- **Game Phase II:** Given the SFL tenants (leaders)' pricing strategies $\{P_{i,j}\}_{i=1}^{M}$, each local device $u_j \in \mathcal{U}$ acts as a Stackelberg follower who chooses its reactive device participation levels $\{q_{i,j}\}_{i=1}^{M}$ for various downstream tasks to maximize its device utility $\lambda_j$, as defined in Problem **P2**.

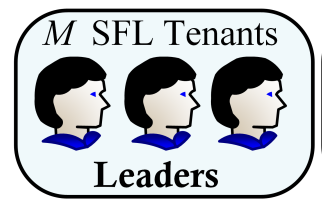

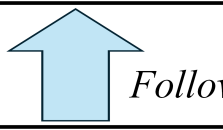

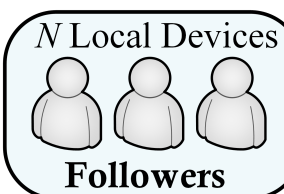

Fig. 2: Multi-leader multi-follower Stackelberg game between SFL tenants and local devices.

We refer to the proposed Stackelberg game as Incentive-driven Device Participation Game (IDP Game). The solution concept of the IDP Game is Stackelberg equilibrium (SE), which we define as follows.

***Definition* 1** (*Solution Concept of the Proposed IDP Game*)**.** The Stackelberg Equilibrium (SE) of the IDP Game is defined by a set of decisions $\{\mathbf{P}_i^*, \mathbf{q}_i^*\}_{i=1}^{M}$ that satisfies the following conditions:

$$q_{i,j}^*(\mathbf{P}_i, \mathbf{P}_{\text{-}i}) = \underset{q_{i,j}}{\operatorname{argmax}}\ \lambda_j(\mathbf{P}_i, \mathbf{P}_{\text{-}i}),\ \forall u_j \in \mathcal{U}, \psi_i \in \Psi \quad (16a)$$

$$\mathbf{P}_i^* = \underset{\mathbf{P}_i}{\operatorname{argmin}}\ \Lambda_i\left(\mathbf{P}_i, \mathbf{q}_i^*(\mathbf{P}_i, \mathbf{P}_{\text{-}i}^*)\right),\ \forall \psi_i \in \Psi \quad (16b)$$

where $\mathbf{q}_i^* = \{q_{i,1}^*, ..., q_{i,N}^*\}$. The local device's decision on the device participation level for downstream task $i$ depends not only on the pricing strategy $\mathbf{P}_i$ of the SFL tenant $\psi_i$ but also on the pricing strategies of competing SFL tenants, denoted by $\mathbf{P}_{\text{-}i} = \{\mathbf{P}_1, ..., \mathbf{P}_{i\text{-}1}, \mathbf{P}_{i+1}, ..., \mathbf{P}_M\}$.

At a SE, neither the SFL tenants nor the local devices would have incentive to deviate from their chosen strategies. A well-established method to obtain the SE is backward induction [26]. This involves first solving for each local device $u_j$'s decision-making $\{q_{i,j}^*\}_{i=1}^{M}$ in Game Phase II, given the SFL tenants' pricing-strategy profile $\{\mathbf{P}_i\}_{i=1}^{M}$. Subsequently, we work backward to Game Phase I to finalize the optimal pricing-strategy profile $\{\mathbf{P}_i^*\}_{i=1}^{M}$ for each SFL tenant $\psi_i \in \Psi$. However, solving the IDP Game poses the following unique research challenges:

- **Challenge 1 (Independent Device Participation):** Each local device $u_j$ independently determines its participation in downstream tasks based on the price incentives offered by various SFL tenants. Such selective participation may lead to certain downstream tasks being dominated by data from a small subset of devices with disproportionately high participation levels, leading to FM model biases for SFL tenants.
- **Challenge 2 (Device Contribution Assessment):** The absence of an analytical formulation for the expected global FM loss $\Lambda_i = \mathbb{E}\left[F\left(\mathbf{w}_i^{\Gamma}(\mathbf{q}_i)\right)\right]$ hinders the SFL tenant's ability to gauge device contributions to FM fine-tuning. Moreover, before completing the FM fine-tuning, it is difficult to predict how device participation levels $\mathbf{q}_i$ impact the final FM $\mathbf{w}_i^{\Gamma}(\mathbf{q}_i)$ and the corresponding loss.
- **Challenge 3 (Inter-Tenant Device Competition):** As local devices are shared among multiple SFL tenants, these SFL tenants naturally compete for robust local devices with high-quality training data. Therefore, effectively managing inter-tenant competition is essential to ensure balanced FM fine-tuning performance across all SFL tenants.

We will respectively handle these three research challenges in the following Sections IV-B through IV-D, and then implement a decentralized price-incentive mechanism to solve our proposed ISCP Game in Section V.

### *B. Bias-Resilient Split Federated Learning*

To tackle the Challenge 1, we propose a bias-resilient SFL scheme that accommodates independent device participation. Specifically, we design an unbiased global FM synchronization procedure for the SFL tenant $\psi_i$'s downstream task, ensuring that our globally synchronized FM model $\mathbf{w}_i(\mathbf{q}_i)$, obtained with independent device participation levels $\mathbf{q}_i = \{q_{i,1}, ..., q_{i,N}\}$, remains unbiased towards full device participation. For any training round $r \in \{kI \,|\, k \in \mathbb{Z}^+\}$ in which global FM synchronization takes place, we define the globally synchronized FM model with full device participation as $\overline{\mathbf{w}}_i^r = \sum_{j=1}^{N} a_{i,j} \mathbf{w}_{i,j}^r$. Meanwhile, we introduce an auxiliary variable $[k]$ to denote the SFL synchronization cycle spanning training rounds $(k\text{-}1)I{+}1$ and $kI$. At the start of the SFL synchronization cycle $[k]$, the global FM model is synchronized, and local devices update their SFL participation decisions for downstream tasks. With these definitions, we have the following theorem.

**Theorem 1** (*Bias-Resilient Global SFL Model Aggregation*). Consider $\mathcal{U}_i^{[k]}$ as the set of local devices participating in the SFL tenant $\psi_i$'s downstream task during the SFL synchronization cycle $[k]$, $\forall k \in \mathbb{Z}^+$. For each participating device $u_j \in \mathcal{U}_i^{[k]}$, its device-side ($\mathbf{w}_{i,j}^{C,r}$) and server-side ($\mathbf{w}_{i,j}^{S,r}$) submodel updates are assembled at the training round $kI$ to forge the device model $\mathbf{w}_{i,j}^{r}$ as:

$$\mathbf{w}_{i,j}^{kI} \leftarrow (\mathbf{w}_{i,j}^{C,kI} \parallel \mathbf{w}_{i,j}^{S,kI}). \quad (17)$$

The forged device models $\mathbf{w}_{i,j}^{kI}$ of all participating local devices $u_j \in \mathcal{U}_i^{[k]}$ are then aggregated to update the global FM model $\mathbf{w}_i^{kI}$ as:

$$\mathbf{w}_i^{kI}(\mathbf{q}_i) \leftarrow \mathbf{w}_i^{(k\text{-}1)I} + \sum_{u_j \in \mathcal{U}_i^{[k]}} \left( \frac{a_{i,j}}{q_{i,j}} \cdot \left( \mathbf{w}_{i,j}^{kI} - \mathbf{w}_i^{(k\text{-}1)I} \right) \right) \quad (18)$$

This global FM synchronization procedure guarantees that:

$$\mathbb{E}_{\mathcal{U}_i^{[k]}}[\mathbf{w}_i^{kI}(\mathbf{q}_i)] = \overline{\mathbf{w}}_i^{kI}. \quad (19)$$

*Proof.* See Appendix A of the supplementary material. □

Algorithm 1 elaborates the bias-resilient SFL process with independent device participation, which modifies the standard FedAvg approach based on Theorem 1. The methodology for determining the optimal device participation levels $\mathbf{q}_i$ will be discussed in Section IV-D. The key differences from the de facto FedAvg method are *independent device participation* (*Lines 1-2*) and *bias-resilient global FM synchronization* (*Lines 21-23*). Specifically,

**Algorithm 1:** Bias-Resilient Split Federated Learning with Independent Device Participation

**Input:** Downstream task of SFL tenant $\psi_i$;
SFL synchronization cycle $[k]$;
The last global FM model $\mathbf{w}_i^{(k\text{-}1)I}$;
Device participation levels $\mathbf{q}_i$.
**Output:** New global FM model $\mathbf{w}_i^{kI}$.

1 /* **Independent Device Participation** */
2 Obtain local devices $\mathcal{U}_i^{[k]} \subseteq \mathcal{U}$ that join SFL tenant $\psi_i$'s downstream task, per their device participation levels $\mathbf{q}_i$;
3 /* **Device-Edge Synergistic FM Fine-tuning** */
4 **for** *each local device* $u_j \in \mathcal{U}_i^{[k]}$ *in parallel* **do**
5 $\quad$ ▷ **Device-side submodel update:**
6 $\quad$ Download $\mathbf{w}_{i,j}^{C,(k\text{-}1)I}$ to the local device $u_j$;
7 $\quad$ **for** *the training round* $r \in [k]$ **do**
8 $\quad\quad$ Forward on $\mathbf{w}_{i,j}^{C,r\text{-}1}$ to get the activation $\mathbf{A}_{i,j}$;
9 $\quad\quad$ Send $\mathbf{A}_{i,j}$ to the edge server $e$ for updating $\mathbf{w}_{i,j}^{S,r\text{-}1}$;
10 $\quad\quad$ Wait until receiving the intermediate gradient $\nabla F_j(\mathbf{w}_{i,j}^{s_{i,j}+1};\, \mathbf{A}_{i,j})$ from the edge server $e$;
11 $\quad\quad$ Update $\widetilde{\mathbf{w}}_{i,j}^{C,r} \leftarrow \widetilde{\mathbf{w}}_{i,j}^{C,r\text{-}1} - \gamma \cdot \nabla F_j(\widetilde{\mathbf{w}}_{i,j}^{C,r\text{-}1}, \mathbf{X}_{i,j})$;
12 $\quad$ Upload $\mathbf{w}_{i,j}^{C,kI}$ to the edge server $e$;
13 $\quad$ ▷ **Server-side submodel update:**
14 $\quad$ Initialize $\mathbf{w}_{i,j}^{S,(k\text{-}1)I}$ on the edge server $e$;
15 $\quad$ **for** *the training round* $r \in [k]$ **do**
16 $\quad\quad$ Wait until receiving $\mathbf{A}_{i,j}$ from the local device $u_j$;
17 $\quad\quad$ Forward propagate $\mathbf{A}_{i,j}$ on $\mathbf{w}_{i,j}^{S,r\text{-}1}$;
18 $\quad\quad$ Calculate loss $F_j(\mathbf{w}_{i,j}^{S,r\text{-}1};\, \mathbf{A}_{i,j})$ and its gradients;
19 $\quad\quad$ Update $\widetilde{\mathbf{w}}_{i,j}^{S,r} \leftarrow \widetilde{\mathbf{w}}_{i,j}^{S,r\text{-}1} - \gamma \cdot \nabla F_j(\widetilde{\mathbf{w}}_{i,j}^{S,r\text{-}1}; \mathbf{A}_{i,j})$;
20 $\quad\quad$ Reply to the local device $u_j$ with $\nabla F_i(\mathbf{w}_{i,j}^{s_{i,j}+1}; \mathbf{A}_{i,j})$;
21 /* **Bias-Resilient Global FM Synchronization** */
22 $\mathbf{w}_{i,j}^{kI} \leftarrow \left(\mathbf{w}_{i,j}^{C,kI} \parallel \mathbf{w}_{i,j}^{S,kI}\right)$;
23 $\mathbf{w}_i^{kI} \leftarrow \boldsymbol{w}_i^{(k-1)I} + \sum_{u_j \in \mathcal{U}_i^{[k]}} \frac{a_{i,j}}{q_{i,j}} \left(\boldsymbol{w}_{i,j}^{kI} - \boldsymbol{w}_i^{(k-1)I}\right)$;
24 **return** $w_i^{kI}$;

- Unlike the active device sampling schemes [27][28] in which device sampling probabilities $q_{i,j}^{\text{sam}}$ are dependent with $\sum_{j=1}^{N} q_{i,j}^{\text{sam}} = 1$, we consider each local device $u_j$'s participation level $q_{i,j}$ in SFL tenant $\psi_i$'s downstream task as mutually independent. This enables greater flexibility for local devices to determine their device participation levels based on self-interested factors, including received monetary incentives and local FM fine-tuning costs.
- Inspired by the adaptive importance sampling [29], we re-weights each participating local device $u_j$'s FM fine-tuning updates $\mathbf{w}_{i,j}$ by the inverse of its device participation level $q_{i,j}$ during the SFL tenant $\psi_i$'s global FM synchronization stage. Intuitively, local devices with low participation levels can still impact the global FM model once they participate in the downstream task. This ensures that the obtained global FM model is unbiased towards that with full device participation.

### C. Convergence Analysis of Split Federated Learning

To tackle the Challenge 2, we derive a rigorous convergence bound for our proposed bias-resilient SFL scheme (i.e., Algorithm 1) to forecast device contributions towards FM fine-tuning. This convergence bound reveals an analytical relationship between $\mathbf{q}_i$ and $\mathbb{E}\left[F\left(\mathbf{w}_i^{\Gamma}\left(\mathbf{q}_i\right)\right)\right]$. As a result, we can approximate $\mathbb{E}\left[F\left(\mathbf{w}_i^{\Gamma}\left(\mathbf{q}_i\right)\right)\right]$ in the SFL tenant's decision problem **P1**, which guides SFL tenants in developing optimal pricing strategies within our incentive mechanism. Before presenting our convergence analysis result, we define $\overline{\mathbf{w}}_i^*$ as the globally optimal FM model obtained under full device participation, as described in Eq. (20). It serves as the optimal baseline for evaluating the learning performance achieved by our proposed bias-resilient SFL scheme (i.e., Algorithm 1).

$$\overline{\mathbf{w}}_i^* = \underset{\mathbf{w}_i}{\arg\min} \sum\nolimits_{u_j \in \mathcal{U}} a_{i,j} \cdot F_j(\mathbf{w}_i; \mathbf{X}_{i,j}). \tag{20}$$

Meanwhile, we preliminarily make the following assumptions, which are typical in the SFL convergence analysis literature [10][30][31].

**Assumption 1** (*Loss Functions*)**.** For each local device $u_j \in \mathcal{U}_i$ participating in the SFL tenant $\psi_i$'s downstream task, its local loss function $F_j$ satisfies the $L_i$*-smoothness* and $\mu_i$*-Polyak-lojasiewicz inequality* conditions:

*1)* $L_i$*-smoothness:*

$F_j(\mathbf{w}_{i,j}) \leq F_j(\mathbf{w}'_{i,j}) + (\mathbf{w}_{i,j} - \mathbf{w}'_{i,j})^T \nabla F_j(\mathbf{w}'_{i,j}) + \frac{L_i}{2} \|\mathbf{w}_{i,j} - \mathbf{w}'_{i,j}\|_2^2$.

*2)* $\mu_i$*-Polyak-Lojasiewicz inequality:*

$F_j(\mathbf{w}_{i,j}) \leq F_j(\overline{\mathbf{w}}_i^*) + \frac{1}{2\mu_i} \|\nabla F_j(\mathbf{w}_{i,j})\|_2^2$.

These inequalities hold for all $\mathbf{w}_{i,j}$ and $\mathbf{w}'_{i,j}$. As a linear combination of local loss functions $F_j$ of all $u_j \in \mathcal{U}_i$, the global loss function $F(\mathbf{w}_i)$ inherits the $L_i$*-smoothness* and $\mu_i$*-Polyak-ojasiewicz inequality* properties.

**Assumption 2** (*Device Gradient Magnitude*)**.** For each local device $u_j \in \mathcal{U}_i$ participating in the SFL tenant $\psi_i$'s downstream task, the expected magnitude of its stochastic gradients $\nabla F_j\left(\mathbf{w}_{i,j}; \mathbf{X}_{i,j}\right)$ is bounded by $G_{i,j}^2$:

$$\mathbb{E}\left\|\nabla F_j\left(\mathbf{w}_{i,j}; \mathbf{X}_{i,j}\right)\right\|^2 \leq G_{i,j}^2.$$

*Remark:* To better reflect the non-i.i.d data distribution across local devices, we hereby refine Assumption 2 from prior studies [10][30][31] by introducing a distinct gradient-magnitude bound $G_{i,j}$ for each local device $u_j \in \mathcal{U}_i$, instead of applying a uniform bound $G_i$ across all participating local devices. This modified assumption enables a more accurate pricing-strategy design for SFL tenants. In practice, $G_{i,j}$ can be estimated by tracking historical local stochastic gradient norms of the participating local devices.

**Assumption 3** (*Intra-Device Gradient Variance*)**.** For each local device $u_j \in \mathcal{U}_i$ participating in the SFL tenant $\psi_i$'s downstream task, the stochastic gradient of its local loss function $F_j$ is unbiased with its variance bounded by $\sigma_{i,j}^2$:

$$\mathbb{E}\left\|\nabla F_j\left(\mathbf{w}_{i,j}; x_{i,j}^k\right) - \nabla F_j\left(\mathbf{w}_j\right)\right\|^2 \leq \sigma_{i,j}^2, \forall\, x_{i,j}^k \in \mathbf{X}_{i,j}$$

where $\nabla F_j\left(\mathbf{w}_{i,j}\right) = \mathbb{E}\left[\nabla F_j\left(\mathbf{w}_{i,j}, x_{i,j}^k\right)\right]$.

**Theorem** 2 (*SFL Convergence Bound with Independent Device Participation*). Let Assumptions 1 to 3 hold, and $L_i$, $\mu_i$, $G_{i,j}$ $\sigma_{i,j}$ be defined therein. Given any device participation level $\mathbf{q}_i = \{q_{i,j}\}_{j=1}^{N}$ and the bias-resilient global SFL model aggregation scheme in Theorem 1, if the decaying learning rate $\gamma_{i,k} = \frac{2}{\max\{8L_i, \mu_i \cdot I\} + \mu_i \cdot k}$ at the SFL synchronization cycle $[k]$, then the *optimality gap* compared with $\overline{\mathbf{w}}_i^*$ after $K_i^{\Gamma}$ SFL synchronization cycles satisfies:

$$\mathbb{E}\left[F\left(\mathbf{w}_i^{\Gamma}\right)\right] - F(\overline{\mathbf{w}}_i^*) \leq \frac{1}{K_i^{\Gamma}} \left( \alpha_i \sum_{j=1}^{N} \frac{(1-q_{i,j}) a_{i,j}^2 G_{i,j}^2}{q_{i,j}} + \beta_i \right), \quad (21)$$

where $\alpha_i = \frac{8L_i I}{\mu_i^2}$, $\beta_i = \frac{2L_i}{\mu_i^2 I} A_0 + \frac{12 L_i^2}{\mu_i^2 I} A_1 + \frac{4L_i^2}{\mu_i I} \left\| \mathbf{w}_i^0 - \mathbf{w}_i^* \right\|^2$, $A_0 = \sum_{j=1}^{N} a_{i,j}^2 \sigma_{i,j}^2 + 8 \sum_{j=1}^{N} a_{i,j} G_{i,j}^2 (I-1)^2$, and $A_1 = F(\overline{\mathbf{w}}_i^*) - \sum_{j=1}^{N} a_{i,j} \cdot \min_{\mathbf{w}_i} F_j(\mathbf{w}_i; \mathbf{X}_{i,j})$.

*Proof.* See Appendix B of the supplementary material. □

Theorem 2 provides us with the following key insights:

- As a key difference from existing convergence analysis results that assume *altruistic* and *obedient* devices [10][30][31], our derived SFL convergence bound in Eq. (21) respects the autonomy of local devices in decision-making. It holds for arbitrary device participating levels $\mathbf{q}_i$, and accommodates a flexible number of participating local devices at each SFL synchronization cycle.
- The SFL convergence bound in Eq. (21) illustrates how stochastic device participation (i.e., $q_{i,j} < 1$) impacts the SFL convergence of a downstream task compared to full device participation, where reduced device participation levels $q_{i,j}$ generally degrade the SFL convergence rate. Meanwhile, it highlights that obtaining an unbiased global foundation model necessitates a non-zero device participation level from each local device, i.e., $q_{i,j} > 0$, $\forall u_j \in \mathcal{U}$. As $q_{i,j} \to 0$, it would take an infinite number of SFL synchronization cycles for FM fine-tuning convergence.
- The SFL convergence bound in Eq. (21) establishes the relationship between the expected global FM loss $\mathbb{E}\left[F\left(\mathbf{w}_i^{\Gamma}\left(\mathbf{q}_i\right)\right)\right]$, the device participation levels $\mathbf{q}_i$, and the local training data's statistical heterogeneity (indicated by $a_{i,j}$ and $G_{i,j}$). In other words, it assesses how local device's unbalanced data ($a_{i,j}$) and non-i.i.d. data distribution ($G_{i,j}$) affect the FM fine-tuning under $\mathbf{q}_i$. This offers an analytical formulation for the expected global FM loss $\mathbb{E}\left[F\left(\mathbf{w}_i^{\Gamma}\left(\mathbf{q}_i\right)\right)\right]$ in the SFL tenant's decision problem **P1**, facilitating an effective pricing-strategy decision.

In summary, we can leverage the SFL convergence bound derived in Eq. (21) as a surrogate to evaluate the global FM loss if selecting different device participation levels $\mathbf{q}_i$. This approach is widely accepted in the distributed machine learning community [19][32][33], as it is typically infeasible to precisely determine how varying distributed machine learning configurations (e.g., different device participation levels $\mathbf{q}_i$ in this paper) will impact the final FM fine-tuning performance before the model training process is fully completed.

### *D. Stackelberg Equilibrium Analysis of IDP Game*

To address Challenge 3, we model inter-tenant device competition as a congestion game, establishing a management framework that ensures fair and balanced opportunities for each SFL tenant to solicit high-quality device participation. Prior to this, we solve our two-phase IDP Game using backward induction. Starting with Game Phase II, we first solve the local devices' reactive decision-making $\{q_{i,j}^*(\mathbf{P}_i, \mathbf{P}_{\text{-}i})\}_{i=1}^{M}$. We then move to Game Phase I to finalize the pricing-strategy profile $\{\mathbf{P}_i^*\}_{i=1}^{M}$ through congestion game theory, ensuring that it reaches the SE.

**Local Devices' Decisions in Game Phase II:** Assuming a pricing-strategy profile $\{\mathbf{P}_i\}_{i=1}^{M}$ set by SFL tenants, each local device $u_j \in \mathcal{U}$ respond to determine its optimal device participation levels $\{q_{i,j}^*(\mathbf{P}_i, \mathbf{P}_{\text{-}i})\}_{i=1}^{M}$ for various downstream tasks by solving Problem **P2**. It is straightforward to observe that Problem **P2** is a convex optimization problem, aligning with the canonical structure of convex optimization (the proof is omitted here for brevity). Accordingly, the solution to Problem **P2** can be obtained by satisfying the Karush-Kuhn-Tucker (KKT) conditions, as formulated below:

$$\frac{\partial \lambda_j}{\partial q_{i,j}^*} + \sum_{c=1}^{2} \left( \mathcal{L}_j^{(c)} \cdot \frac{\partial \phi_j^{(c)}}{\partial q_{i,j}^*} \right) = 0, \qquad \forall i \in \{1, ..., M\} \quad (22)$$

$$\mathcal{L}_j^{(c)} \cdot \phi_j^{(c)}(\{q_{i,j}^*\}_{i=1}^{M}) = 0, \qquad \forall c \in \{1, 2\} \quad (23)$$

$$\phi_j^{(c)}(\{q_{i,j}^*\}_{i=1}^{M}) \geq 0, \qquad \forall c \in \{1, 2\} \quad (24)$$

$$\mathcal{L}_j^{(c)} \geq 0, \qquad \forall c \in \{1, 2\} \quad (25)$$

where $\mathcal{L}_j^{(c)}$, $\forall c \in \{1, 2\}$ are the Lagrange multipliers, and the inequality constraint functions $\phi_j^{(c)}(\{q_{i,j}^*\}_{i=1}^{M})$, $\forall c \in \{1, 2\}$ arise from the constraints (15a) in **P2**, defined as:

$$\phi_j^{(1)} = 1 - \sum\nolimits_{i=1}^{M} q_{i,j}, \quad \phi_j^{(2)} = \sum\nolimits_{i=1}^{M} q_{i,j} \quad (26)$$

**SFL Tenants' Decisions in Game Phase I:** Each SFL tenant $\psi_i \in \Psi$ determines its optimal pricing strategy $\mathbf{P}_i^*$, based on the local devices' decision responses $\{q_{i,j}^*(\mathbf{P}_i, \mathbf{P}_{\text{-}i})\}_{i=1}^{M}$ obtained in Game Phase II. Mathematically, the SFL tenant $\psi_i$ substitutes the derived numerical solution $\{q_{i,j}^*(\mathbf{P}_i, \mathbf{P}_{\text{-}i})\}_{i=1}^{M}$ into the optimization objective function $\Lambda_i$ in Problem **P1** to calculate the optimal price vector $\mathbf{P}_i$ under the budget constraint $B_i$.

However, given the considerable challenge of deriving an analytical formulation for the Problem **P1**'s optimization objective $\Lambda_i = \mathbb{E}\left[F\left(\mathbf{w}_i^{\Gamma}\left(\mathbf{q}_i\right)\right)\right]$, we first approximate it with our obtained SFL convergence bound in Eq. (21). In addition, the Problem **P1**'s decision vector $\mathbf{P}_i$ exclusively controls the term $\alpha_i / K_i^{\Gamma} \cdot \sum_{j=1}^{N} \frac{(1-q_{i,j}) a_{i,j}^2 G_{i,j}^2}{q_{i,j}}$ in the SFL convergence bound. Consequently, the SFL tenant's decision problem **P1** can be formulated as the following Problem **P1**$'$:

$$(\mathbf{P1}'): \min_{\mathbf{P}_i} \ \Lambda_i' = \frac{\alpha_i}{K_i^{\Gamma}} \sum_{j=1}^{N} \frac{\left(1 - q_{i,j}^*(\mathbf{P}_i, \mathbf{P}_{\text{-}i})\right) a_{i,j}^2 G_{i,j}^2}{q_{i,j}^*(\mathbf{P}_i, \mathbf{P}_{\text{-}i})} \quad (27)$$

$$s.t. \quad \text{Eq. (13a)}$$

Owing to inter-tenant competition, the local device's decision response $q_{i,j}^*(\mathbf{P}_i, \mathbf{P}_{\text{-}i})$ in **P1**$'$ depends not only on the pricing strategy $\mathbf{P}_i$ of SFL tenant $\psi_i$ itself, but is also influenced by the pricing strategies $\mathbf{P}_{\text{-}i}$ of other SFL tenants. The local device $u_j$ compares the price benefits offered by various

SFL tenants, and reflects its preference in $q^*_{i,j}$, prioritizing participation in the SFL tenant's downstream task that provides greater monetary incentives. On the other hand, each SFL tenant $\psi_i$ is self-motivated to secure a competitive advantage by optimizing its pricing strategy in Problem $\mathbf{P1}'$. To analyze how inter-tenant device competition informs the SFL tenants of their pricing-strategy decisions, we model the interactions amongst multiple SFL tenants as a congestion game $\Omega$.

***Definition* 2** (*Congestion Game*)**.** The inter-tenant competition for the shared local devices $\mathcal{U}$ is formulated as *a multi-SFL-tenant congestion game* $\Omega = \langle \Psi, \{\mathbf{P}_i\}_{i=1}^M, \{\Lambda'_i\}_{i=1}^M \rangle$, where:

- $\Psi$ is the set of players, specifically $M$ SFL tenants here. These players $\psi_i \in \Psi$ compete to achieve a lower global FM loss via incentivizing high-quality SFL participation from local devices.
- $\{\mathbf{P}_i\}_{i=1}^M$ is the strategy profile, where each $\mathbf{P}_i$ corresponds to the SFL tenant $\psi_i$'s pricing strategy. $\mathbf{P}_i = \{P_{i,1}, ..., P_{i,N}\}$ specifies the customized price $P_{i,j}$ for each local device $u_j \in \mathcal{U}$ participating in the downstream task of SFL tenant $\psi_i$.
- $\Lambda'_i$ is the disutility function of SFL tenant $\psi_i$, formulated by Eq. (27), which indicates the global FM loss gained by SFL tenant $\psi_i$ via adopting a pricing strategy $\mathbf{P}_i$.

Each SFL tenant $\psi_i \in \Psi$ aims to incentivize robust local devices participating in its downstream task, which aids in minimizing its global FM loss. Suppose that a SFL tenant $\psi_i$ initially selects a pricing strategy $\mathbf{P}_i$ but finds an alternative budget-feasible strategy $\mathbf{P}'_i$ that yields a lower global FM loss (indicated by $\Lambda_i(\mathbf{P}'_i)$), the SFL tenant $\psi_i$ would naturally seek to improve its pricing strategy as $\mathbf{P}'_i$. The strategy-improvement procedure moves forward iteratively, with each iteration allowing only one SFL tenant to win the opportunity to adjust its pricing strategy. The congestion game iterates until no SFL tenant can improve its global model loss by altering its game strategy, finalizing the pricing-strategy profile $\{\mathbf{P}^*_i\}_{i=1}^M$ for all SFL tenants, where a Nash Equilibrium (NE) is reached.

***Definition* 3** (*Nash Equilibrium*)**.** A Nash Equilibrium for the multi-SFL-tenant congestion game $\Omega = \langle \Psi, \{\mathbf{P}_i\}_{i=1}^M, \{\Lambda'_i\}_{i=1}^M \rangle$ is a pricing-strategy profile $\{\mathbf{P}^*_i\}_{i=1}^M$ satisfying that for each SFL-tenant player $\psi_i \in \Psi$,

$$\Lambda'_i(\mathbf{P}^*_i, \mathbf{P}^*_{\text{-}i}) \geq \Lambda'_i(\mathbf{P}_i, \mathbf{P}^*_{\text{-}i}). \tag{28}$$

In essence, the multi-SFL-tenant congestion game $\Omega$ acts as a Phase-I subgame within our two-phase IDP Game. After specifying the local devices' Phase-II decision responses $q^*_{i,j}(\mathbf{P}_i, \mathbf{P}_{\text{-}i})$, solving the SE for our IDP Game can be simplified through backward induction. This reduces the problem to finding the NE solution $\{\mathbf{P}^*_i\}_{i=1}^M$ of the Phase-I congestion game $\Omega$. In other words, the IDP Game naturally reaches its SE when the congestion game $\Omega$ converges to a NE.

## V. Decentralized Price-Incentive Algorithm

### A. Decentralized Algorithm Implementation

We implement a decentralized Price-Incentive Algorithm (PRINCE) for multi-tenant SFL, enabling the practical implementation of our two-phase IDP Game to find the SE solution as detailed in Section IV-D. As independent entities, both SFL tenants and local devices are accommodated to make independent decisions based on their own interests, until a SE is reached.

**Algorithm 2:** Decentralized Price-Incentive Algorithm for Multi-Tenant Split Federated Learning (PRINCE)

1 **Initialize** the pricing-strategy profile $\{\mathbf{P}_i\}_{i=1}^M$;
2 `/* Two-Phase IDP Game */`
3 **repeat**
4     ▷ `Local devices' decisions in Game Phase II :`
5     **for** *each local device* $u_j \in \mathcal{U}$ *in parallel* **do**
6         Calculate $\{q^*_{i,j}(\mathbf{P}_i,\mathbf{P}_{\text{-}i})\}_{i=1}^M$ that maximizes its device utility $\lambda_j$ by solving the Problem $\mathbf{P2}$;
7     ▷ `SFL tenants' decisions in Game Phase I :`
8     **for** *each SFL tenant* $\psi_i \in \Psi$ *in parallel* **do**
9         Calculate $\Lambda'_i$ based on the reactive SFL participation levels $q^*_{i,j}(\mathbf{P}_i,\mathbf{P}_{\text{-}i})$ of all local devices $u_j$;
10         **if** $\exists\, \mathbf{P}'_i \neq \mathbf{P}_i$, $\Lambda'_i(\mathbf{P}'_i,\mathbf{P}_{\text{-}i}) < \Lambda'_i(\mathbf{P}_i,\mathbf{P}_{\text{-}i})$ *s.t. Eq.(13a)* **then**
11             Propose a strategy-change request $\mathbf{P}'_i$;
12     Select the winner $\psi_w$ whose strategy improvement $\mathbf{P}'_w$ is approved for yielding the greatest improvement in $\sum_{i=1}^M \Lambda'_i$;
13     Update the winner $\psi_w$'s pricing strategy $\mathbf{P}_w \leftarrow \mathbf{P}'_w$;
14 **until** *no SFL tenant is approved to adjust its pricing strategy*;
15 The IDP Game reaches a SE in which $\mathbf{P}^*_i \leftarrow \mathbf{P}_i$, $\forall\, \psi_i \in \Psi$;
16 **return** the SE solution $\{\mathbf{P}^*_i\}_{i=1}^M$;

The pseudo-code is presented in Algorithm 2. This decentralized algorithm begins with an initial pricing-strategy $\{\mathbf{P}_i\}_{i=1}^M$, where each SFL tenant $\psi_i \in \Psi$ is assigned a uniform pricing strategy $\mathbf{P}_i = \{\frac{B_i}{N}\}_{j=1}^N$ which adheres to the budget constraint in Eq. (13a) (*Line 1*). Next, each local device $u_j \in \mathcal{U}$ concurrently calculates its reactive SFL participation levels $q^*_{i,j}$ under $(\mathbf{P}_i,\mathbf{P}_{\text{-}i})$ (*Lines 4-6*). Informed by the local devices' decision responses, all SFL tenants $\psi_i \in \Psi$ then concurrently calculate their $\Lambda'_i$ that indicates the expected global FM loss under $\{q^*_{i,j}\}_{j=1}^N$, and propose their strategy-change requests $\mathbf{P}'_i$ if applicable (*Lines 7-11*). In each strategy-improvement iteration, multiple SFL tenants successively propose their strategy-change requests. Only one SFL tenant $\psi_w$ is selected to win the opportunity for adjusting its pricing strategy (*Lines 12-13*), as it yields the greatest improvement in $\sum_{i=1}^M \Lambda'_i$. These strategy-improvement iterations continue until no SFL tenant is approved to adjust its pricing strategy, and the algorithm terminates (*Line 14*). Our proposed IDP Game has converged to the SE solution $\{\mathbf{P}^*_i\}_{i=1}^M$, as the output of the decentralized algorithm (*Lines 15-16*).

### B. Algorithm Analysis

**1) Convergence Analysis:** We theoretically demonstrate that our proposed PRINCE algorithm can converge to the SE solution within a finite number of strategy-improvement iterations. Since the PRINCE algorithm is the decentralized implementation of IDP Game, it suffices to justify the finite-time convergence of the two-phase IDP Game. Specifically, we first prove the Phase-I subgame—the multi-SFL-tenant congestion game $\Omega$—is a potential game with a potential function $\Upsilon$. Leveraging the *Finite Improvement Property* of potential games [34], our congestion game $\Omega$ is guaranteed to converge to an NE in finite iterations by searching for the optimum of potential function $\Upsilon$. As a result, the finite-time convergence of the IDP Game is justified, as the converged NE in the Phase-I congestion game $\Omega$ corresponds to the SE solution of the two-phase IDP Game. Based on this analysis framework, we begin by defining potential games as follows.

***Definition* 4** (*Potential Game*)**.** Let $\mathcal{P}$ denote the set of feasible pricing-strategy profiles $\{\mathbf{P}_i\}_{i=1}^{M}$. The game $\Omega$ is considered a potential game if there exists a potential function $\Upsilon:\mathcal{P}\rightarrow\mathbb{R}$, such that for each player $\psi_i\in\Psi$, $\Upsilon(\mathbf{P}')<\Upsilon(\mathbf{P})$ holds for any strategy improvement from $\mathbf{P}=(\mathbf{P}_i,\mathbf{P}_{\text{-}i})$ to $\mathbf{P}'=(\mathbf{P}'_i,\mathbf{P}_{\text{-}i})$ satisfying $\Lambda'_i(\mathbf{P}'_i,\mathbf{P}_{\text{-}i})<\Lambda'_i(\mathbf{P}_i,\mathbf{P}_{\text{-}i})$, where $\mathbf{P},\mathbf{P}'\in\mathcal{P}$.

By Definition 4, the potential function $\Upsilon$ monotonically increases with each strategy improvement until it reaches its optimum, representing the SE convergence. Accordingly, we constructively demonstrate in Theorem 3 that the multi-SFL-tenant congestion game $\Omega$ is a potential game, as shown below:

**THEOREM** 3 (*Multi-SFL-Tenant Potential Game*). The multi-SFL-tenant congestion game $\Omega$ is a potential game, with the potential function $\Upsilon:\mathcal{P}\rightarrow\mathbb{R}$ defined as follows:

$$\Upsilon(\mathbf{P})=\sum\nolimits_{i=1}^{M}\Lambda'_i(\mathbf{P}_i,\mathbf{P}_{\text{-}i}). \tag{29}$$

*Proof.* See Appendix C of the supplementary material. □

Having established that the congestion game $\Omega$ is a potential game, we ensure that the PRINCE algorithm converges to an SE solution within a finite number of strategy-improvement iterations.

**2) Performance Analysis:** We analyze the performance of our PRINCE algorithm in terms of its optimality, specifically assessing how effectively it minimizes the global model loss for multiple SFL tenants $\psi_i\in\Psi$. In general, multi-leader multi-follower Stackelberg games can exhibit non-unique SE solutions [34]. Therefore, it is important to investigate whether the PRINCE mechanism reaches the SE solution that minimizes the global model loss across multiple SFL tenants. Theorem 4 formally establishes the algorithmic optimality as follows:

**THEOREM** 4. The PRINCE algorithm identifies the optimal pricing-strategy profile $\{\mathbf{P}_i^*\}_{i=1}^{M}$, where the total SFL convergence bound $\sum_{i=1}^{M}\Lambda'_i$ is minimized.

*Proof.* See Appendix D of the supplementary material. □

Since $\Lambda'_i$ represents the global FM model loss obtained by SFL tenant $\psi_i$, Theorem 4 informs that the PRINCE algorithm achieves robust optimality by minimizing the total global FM model loss across multiple SFL tenants.

**3) Complexity and Scalability Analysis:** In large-scale system scenarios involving numerous SFL tenants and local devices, the decentralized PRINCE algorithm exhibits strong scalability and high implementation efficiency. According to the algorithmic convergence and performance analyses presented earlier, the SFL tenants and local devices negotiate over a finite number of strategy-improvement iterations to reach the optimal SE solution $\{\mathbf{P}_i^*\}_{i=1}^{M}$. Let $\mathcal{X}$ denote the number of iterations required to achieve the optimal SE solution.

Furthermore, each strategy-improvement iteration incurs a manageable time cost. Let $\triangle t$ represent the communication delay for wireless message exchanges between SFL tenants (at the edge serverv $e$) and local devices. Each SFL tenant first broadcast its latest pricing-strategy profiles to local devices in parallel, which takes $\triangle t$. Then, each local device $u_j$ receives the SFL tenants' pricing-strategy profiles and calculates its reactive SFL participation levels $\{q_{i,j}^*\}_{i=1}^{M}$. Subsequently, the local devices reply their SFL participation levels back to the corresponding SFL tenants in parallel, requiring another $\triangle t$. Upon receiving the responses, the SFL tenants compute $\Lambda'_i$ in parallel and propose strategy improvements $\mathbf{P}'_w$. If an SFL tenant's strategy improvement proposal $\mathbf{P}'_w$ is approved, it takes additional $\triangle t$ to notify local devices, marking the start of the next strategy-improvement iteration.

Wireless communication between SFL tenants and local devices causes noticeable delay, compared to which the computation time at both local devices and SFL tenants for calculating $\{q_{i,j}^*\}_{i=1}^{M}$ or $\Lambda'_i$, either through simple linear porgramming or numerical multiplications, is negligible. Therefore, the overall execution time of PRINCE algorithm can be approximated by $2\cdot\mathcal{X}\cdot\triangle t$. Owing to its decentralized design, the algorithmic execution efficiency remains orthogonal to system scale, meaning it does not grow with the number of participating SFL tenants or local devices.

## VI. PERFORMANCE EVALUATION

### A. Experimental Setup

**1) FM Fine-tuning Workloads and Datasets:** We adopt four representative types of downstream tasks involving different data modalities, including image, text, and audio, as the FM fine-tuning workloads evaluated in our simulation experiments. Each downstream task is managed by a separate SFL tenant, who chooses an appropriate FM model and FM fine-tuning method (e.g., full-parameter fine-tuning, adapter, or LoRA). The details of each type of SFL tenant are as follows:

- **SFL Tenant 1 (Image Classification):** The ViT-B/16 model [35], with 86M parameters, undergoes full-parameter fine-tuning for food image classification using 3,000 real-world images from the Food-4 dataset [36], which includes 4 classes of pizza, risotto, steak, and sushi. These image samples are distributed among local devices in an *unbalanced* (following the power-law distribution) and *non-i.i.d.* (each device randomly contains 1-4 classes) fashion.
- **SFL Tenant 2 (Sentiment Analysis):** The BERT-base model [1], with 110M parameters, undergoes full-parameter fine-tuning for sentence sentiment analysis using 10,657 authentic natural language sentences from the CoLA dataset [37]. These data samples are distributed among local devices following an *unbalanced* power-law distribution.
- **SFL Tenant 3 (Speech to Text):** The Whisper-base model [38], with 74M parameters, undergoes adapter fine-tuning (i.e., fine-tuning the Whisper decoder modules while freezing the others) for Turkish speech-to-text conversion using 16,900 real voice recordings from the Common Voice 8.0 dataset [39]. These audio samples are distributed among local devices in an *unbalanced* manner, following a power-law distribution, and a *non-i.i.d.* fashion, where each device randomly selects either female or male voices.

- **SFL Tenant 4 (Questioning Answering):** The LLaMA 2-7B model [2] undergoes LoRA fine-tuning for question answering using 3,000 dialogue prompts from the DialogSum-3K dataset [40]. These prompt samples are distributed among local devices in an *unbalanced* (following the power-law distribution) manner. To fit the billion-sized LLaMA model into runtime memory, we quantize it to the INT4 format using OPTQ [41].

Each SFL tenant $\psi_i$ employs the optimal model splitting strategy in [12] to minimize the SFL time cost per training round. The FM fine-tuning deadline $\Gamma_i$ and the payment budget $B_i$ are proportional to the complexity of downstream task, including factors such as the FM scale and the difficulty in achieving the target accuracy. The hyperparameters $\alpha_i$ and $G_{i,j}$ associated with the SFL convergence bound $\Lambda_i$ can be estimated, following a similar approach as [28].

**2) Local Devices:** The computation capacity $\xi_j^C$ of each local device $u_j$ is drawn from a uniform distribution over [1567, 3100] GFLOPS, which aligns with the performance levels of typical AIoT devices ranging from NVIDIA Jetson Orin NX to Qualcomm Snapdragon SA8295P SIP. The local FM fine-tuning cost parameter $c_i$ of each device depends on its energy consumption, which ranges between 20 and 40 Watts. Each local device holds training data that is simultaneously relevant to multiple SFL tenants' downstream tasks, i.e., SFL tenants $\psi_i$, $\forall i \in \{1, 2, 3, 4\}$. Accordingly, multiple SFL tenants compete with one another to solicit high-quality device participation for efficient FM fine-tuning. This setting reflects the growing proliferation of real-world intelligent applications where diverse downstream tasks (e.g., Twitter and Instagram [14]) may rely on overlapping training data sources.

**3) Edge Server $e$:** The edge server $e$ is configured with a total computation capacity of 330.32 TFLOPS, equivalent to a GPU server equipped with four NVIDIA GeForce RTX 4090 GPU cards. The computation capacity $\xi_i^S$ allocated to each SFL tenant $\psi_i$ is evenly divided from the edge server's total capacity.

**4) Wireless Networks:** We consider heterogenous wireless communication environments between the edge server and local devices, including 4G LTE-Advanced, 5G, and WiFi 5. Consequently, the wireless data download rate $\zeta_j^{\mathrm{d}}$ between edge server $e$ and local device $u_j$ ranges from 50 to 250 Mbps, while the wireless data upload rate $\zeta_j^{\mathrm{u}}$ is uniformly distributed between 17 and 83 Mbps.

### *B. Performance Benchmarks*

Our proposed PRINCE mechanism is compared against three representative performance benchmarks, including the following two state-of-the-art price-incentive schemes (i.e., FAIR [16] and MSDA [12]):

- **FAIR** [16]**:** Each SFL tenant $\psi_i$ expects to solicit device participation through offering incentives to devices with high-quality training data. Instead of leveraging congestion game modeling $\Omega$, the SFL tenant $\psi_i$ simply implements a weighted pricing strategy $\mathbf{P}_i = \{P_{i,1}, ..., P_{i,N}\}$, where the local device $u_j$'s price $P_{i,j}$ is proportional to its learning quality (e.g., data quantity $|a_{i,j}|$).
- **MSDA** [12]**:** Each SFL tenant $\psi_i$ primarily focuses on optimizing its FM model splitting strategy to minimize the SFL time cost per training round, without any specific incentive design. A fixed (uniform) pricing strategy is simply adopted, where the price $P_{i,j}$ for each local device $u_j$ is set equally by dividing the SFL tenant's payment budget $B_i$ fairly, i.e., $P_{i,j} = B_i/N, \forall\, u_j \in \mathcal{U}$.

In addition, we compare our proposed PRINCE mechanism based on split federated FM fine-tuning against a state-of-the-art privacy-preserving FM fine-tuning approach:

- **FedPEFT** [42]**:** This approach adopts classical FL to enable privacy-preserving FM fine-tuning in edge networks, where the entire FM model is offloaded to local devices without model splitting. The same price-incentive strategy as in our PRINCE mechanism is applied to encourage device participation in FL-based FM fine-tuning tasks.

Lastly, we ablate the bias-resilient global SFL model aggregation component from PRINCE to examine how its removal affects performance. We refer to this variant as **PRINCE w/o bias-resilient aggregation**.

### *C. Experimental Results*

**Bias-Resilient Split Federated Learning:** We first empirically validate the effectiveness of our bias-resilient global SFL model aggregation scheme across various SFL tenants' FM fine-tuning workloads. Figure 3 compares FM fine-tuning performance under varying independent device participation probabilities ($q_{i,j}$ = 0.25, 0.5, and 0.75) with the full device participation scenario ($q_{i,j}$ = 1.0). The results demonstrate that, owing to the effectiveness of our bias-resilient global SFL model aggregation scheme, partial device participation ($q_{i,j}$<1.0) in the downstream tasks of SFL Tenants $1 \sim 3$ achieves FM fine-tuning performance comparable to that of full device participation. Even more notably, the question-answering task (SFL Tenant 4) not only effectively eliminates FM model bias caused by partial device participation, achieving competitive FM fine-tuning performance, but also significantly accelerates FM fine-tuning convergence compared to full device participation. This is because full device participation includes straggling local devices that perform device-side FM fine-tuning slowly, hence delaying the global FM fine-tuning process. The straggler effect is particularly pronounced in the question-answering task (SFL Tenant 4), which requires significantly more local computation resources to compete device-side FM fine-tuning compared to the downstream tasks of other SFL tenants.

**FM Fine-Tuning Runtime Performance:** We simulate a multi-tenant SFL environment at the network edge with four SFL tenants (SFL Tenants $1 \sim 4$) and 100 local devices. The FM fine-tuning runtime performance of these SFL tenants is evaluated under our PRINCE mechanism and other benchmark approaches. Figure 4 illustrates the FM fine-tuning performance across various SFL tenants' downstream tasks.

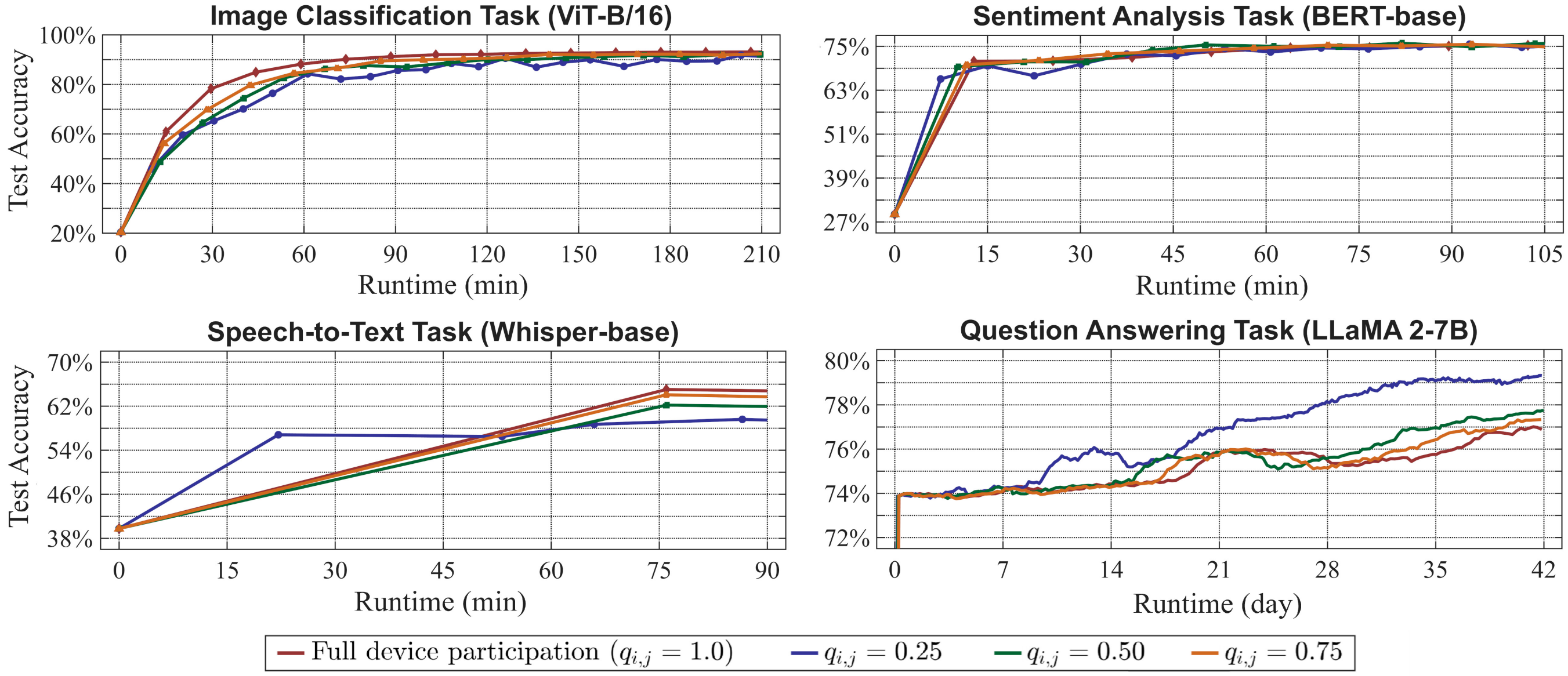

Fig. 3: FM Fine-Tuning Performance across various Independent Device Participation Levels $q_{i,j}$.

Table II compares the FM fine-tuning runtime of different benchmark approaches to reach the target accuracy for various SFL tenants.

Compared to state-of-the-art benchmarks, our proposed PRINCE mechanism consistently achieves the greatest acceleration in FM fine-tuning runtime. Specifically, our PRINCE mechanism achieves the target accuracy, on average, 31.36% faster across various downstream tasks compared to FedPEFT. This observation highlights the superior efficiency of the SFL paradigm over classical FL in handling computation-intensive FM fine-tuning workloads at the network edge. In SFL, the assistance of edge servers enables resource-constrained local devices to contribute to FM fine-tuning. By contrast, in classical FL, these devices often act as performance bottlenecks due to the FM fine-tuning deadline. Consequently, they are excluded from federated FM fine-tuning, which leads to underrepresented device participation and hinders global FM convergence. Meanwhile, the ablation results of PINCE w/o bias-resilient aggregation indicate that the proposed bias-resilient global SFL model aggregation component plays a crucial role in ensuring high-quality global FM convergence, enabling PRINCE to efficiently reach the target accuracy for diverse downstream tasks.

Furthermore, our PRINCE mechanism employs more fine-grained price-incentive strategies, thereby achieving average time savings of 19.31% and 29.97% to reach the target accuracy across various downstream tasks compared to the state-of-the-art price-incentive schemes FAIR and MSDA, respectively. The fixed (uniform) pricing strategy of MSDA treats all local devices equally, failing to adapt price incentives to heterogeneous device contributions toward FM fine-tuning improvements. This limitation hinders its ability to provide targeted incentives to robust local devices, thereby reducing high-quality device participation. The weighted pricing strategy of FAIR accounts for the heterogeneous learning quality of different local devices, but overlooks the impact of inter-tenant competition due to the absence of multi-SFL-tenant congestion game modeling. To secure SFL participation from a high-demand local device, an SFL tenant should offer more competitive pricing incentives than its peers. Our proposed PRINCE mechanism adopts price-incentive strategies that address not only the heterogeneous contributions of devices to FM fine-tuning performance but also the impact of inter-tenant competition. This comprehensive approach enables PRINCE to outperform the state-of-the-art price-incentive schemes, FAIR and MSDA.

**Inter-Tenant Balanced Optimization:** Our PRINCE mechanism optimizes FM fine-tuning performance across multiple SFL tenants in a balanced manner. It effectively coordinates self-interested local devices to participate in various SFL tenants' downstream tasks, thereby meeting each tenant's FM fine-tuning requirements. Table II demonstrates that PRINCE achieves the greatest acceleration in FM fine-tuning runtime across different SFL tenants. This is attributed to our incentive mechanism design, based on the multi-leader multi-follower Stackelberg game, which ensures each SFL tenant has fair opportunities to adjust their pricing strategy, enabling fair competition for device participation in their respective downstream tasks. In contrast, the state-of-the-art price-incentive schemes, FAIR and MSDA, fail to achieve balanced optimization across multiple SFL tenants. FAIR cannot reach the target accuracy of 79% for the question-answering task (SFL Tenant 4) within a reasonable FM fine-tuning deadline, while MSDA fails to reach the target accuracy of 65% for the speech-to-text task (SFL Tenant 3). The absence of the bias-resilient global SFL model aggregation component causes PRINCE w/o bias-resilient aggregation to fail in reaching the target accuracy for all SFL tenants' downstream tasks.

**Impact of Multi-Tenant SFL System Scales:** We also evaluate the efficacy of our PRINCE mechanism across different system scales, considering varying numbers of involved SFL tenants and local devices. Figure 5 compares the total SFL convergence bound, and the total device utility achieved by our PRINCE mechanism against the state-of-the-art price-

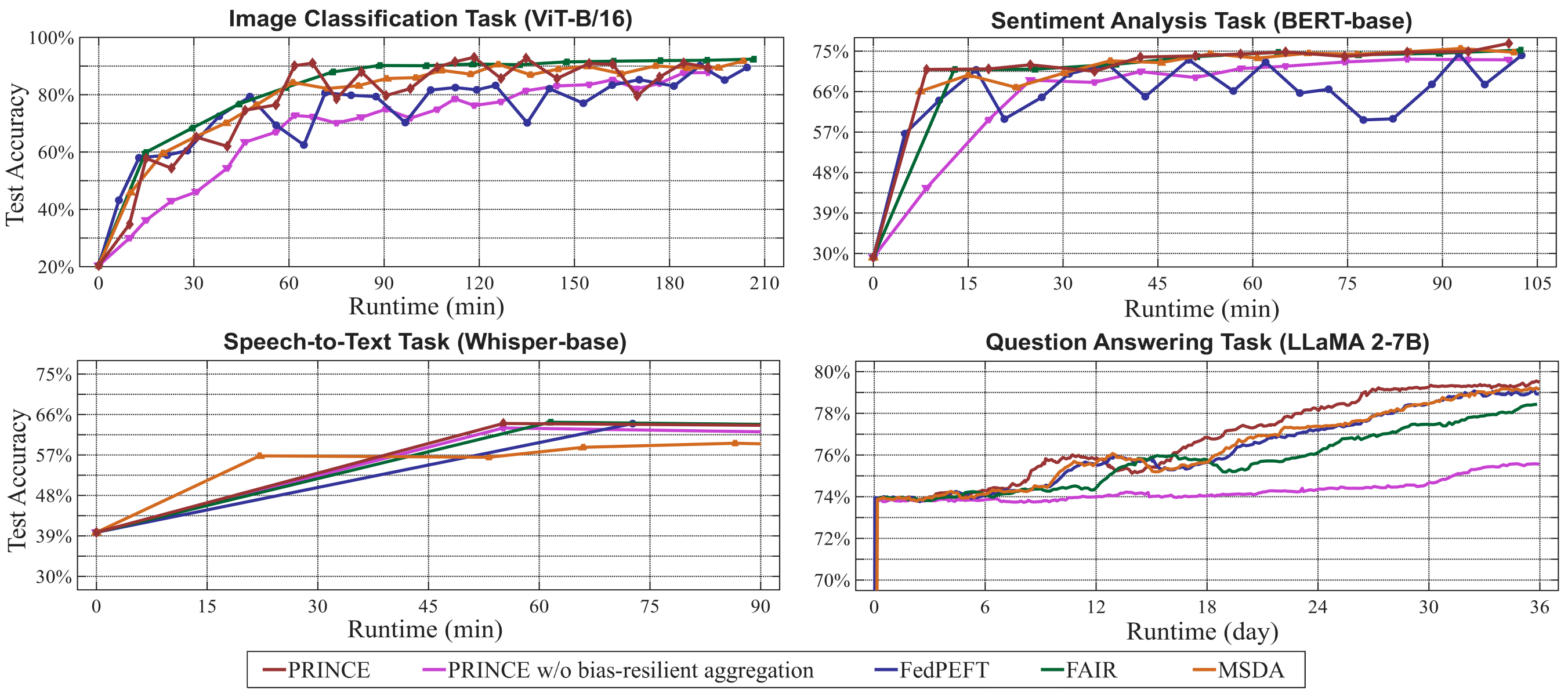

Fig. 4: FM Fine-Tuning Performance across various SFL Tenants' Downstream Tasks at the Network Edge.

TABLE II: FM Fine-Tuning Runtime across various SFL Tenants at the Network Edge (referenced in Fig. 4)

| Methods | SFL Tenants | | | |
|---|---|---|---|---|
| | Image Classification (target accuracy = 90%) | Sentiment Analysis (target accuracy = 73%) | Speech to Text (target accuracy = 65%) | Question Answering (target accuracy = 79%) |
| **PRINCE (ours)** | **61.74 mins** | **42.24 mins** | **110.27 mins** | **26.57 days** |
| PRINCE w/o bias-resilient aggregation | N/A* | N/A* | N/A* | N/A* |
| FedPEFT | 189.61 mins | 49.85 mins | 145.32 mins | 32.64 days |
| FAIR | 88.46 mins | 51.14 mins | 122.93 mins | N/A* |
| MSDA | 126.01 mins | 53.32 mins | N/A* | 32.45 days |

* N/A: Unable to reach the target accuracy within a reasonable FM fine-tuning deadline.

incentive schemes, FAIR and MSDA, at different system scales. The lower the total SFL convergence bound, $\sum_{i=1}^{M} \Lambda_i'$, the smaller the estimated global FM loss achieved across multiple SFL tenants within the FM fine-tuning deadline. We can observe in Figure 5 that our PRINCE mechanism always obtains the lowest total SFL convergence bound, indicating the best FM fine-tuning runtime performance. This aligns with the FM fine-tuning runtime results presented in Figure 3. On the other hand, total device utility reflects the revenue of local devices from SFL participation after deducting local FM fine-tuning costs. While the FAIR incentive scheme achieves the highest total device utility gains, it fails to translate these gains into effective FM fine-tuning performance. This indicates the limited effectiveness of FAIR in multi-tenant SFL environments. Our PRINCE mechanism achieves similar total device utility gains to MSDA, but significantly accelerates the FM fine-tuning process, demonstrating the efficacy of PRINCE.

**Algorithmic Efficiency of our PRINCE Mechanism:** We assess the algorithmic efficiency of our PRINCE mechanism across various multi-tenant SFL system scales, as illustrated in Figure 6. In Algorithm 2, our PRINCE mechanism iteratively adjusts the pricing-strategy profile of multiple SFL tenants, until reaching the SE convergence. Therefore, we use the number of algorithmic iterations required for obtaining the final SE solution as the evaluation metric for algorithmic efficiency. As shown in Figure 6, no remarkable increase trend is observed in the number of required iterations, with varying numbers of involved SFL tenants and local devices. On average, 42 iterations are required to achieve the final SE solution across various system scales, demonstrating the finite-time complexity of our PRINCE mechanism.

**Impact of the SFL Tenant's Payment Budget $B_i$ and Fine-Tuning Deadline $\Gamma_i$:** We evaluate how PRINCE performs under varying payment budgets $B_i$ and FM fine-tuning deadlines $\Gamma_i$ of the SFL tenant. In the multi-tenant competitive environment, we evaluate the impact of $B_i$ and $\Gamma_i$ by adjusting one SFL tenant's parameters $B_i$ and $\Gamma_i$ while keeping others fixed, to facilitate the analysis of PRINCE's performance under different competition conditions. Accordingly, in Figure 7, we vary the SFL tenant 1's payment budget $B_1$ and FM fine-tuning deadline $\Gamma_1$ to measure its individual SFL convergence bound, as well as the total convergence bound of SFL tenants $1 \sim 4$ achieved by PRINCE. With a higher payment budget, the SFL tenant 1 can offer greater incentives to local devices, enhancing device participation in FM fine-tuning. As a result, it gains a competitive advantage and has the potential to attain better global SFL convergence, as evidenced by a lower SFL convergence bound. Similarly, the SFL tenant 1 can extend its FM fine-tuning deadline to allow more time for improving its model fine-tuning performance, reflected in a lower SFL convergence bound. Under varying payment budgets and fine-tuning deadlines of SFL tenant 1, which represent dynamic multi-tenant competition conditions, PRINCE consistently maintains a controllable total SFL convergence bound, demonstrating its robustness in managing and balancing inter-tenant competition.

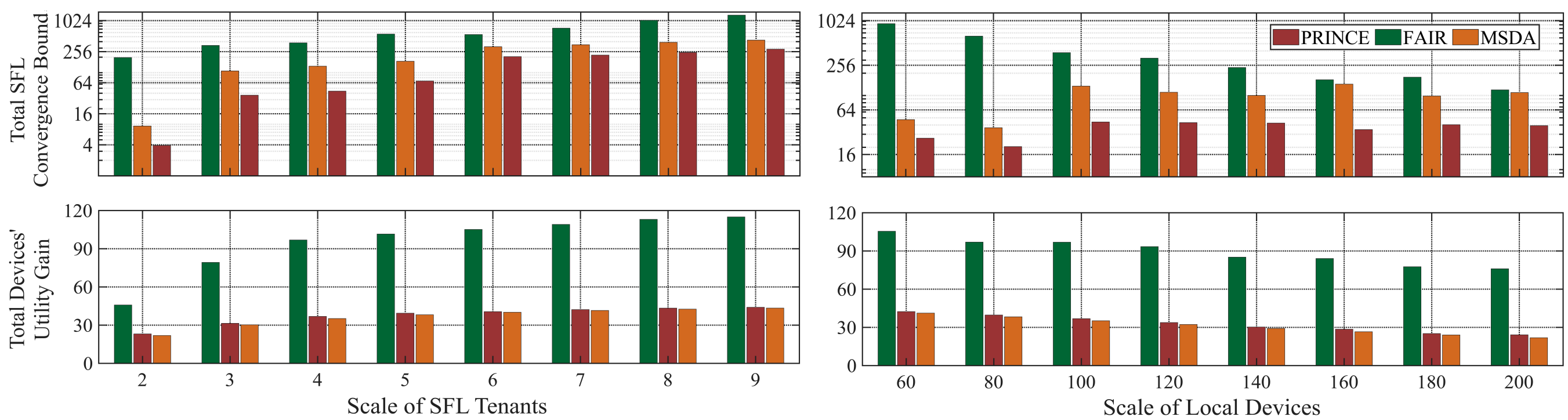

Fig. 5: Impact of Multi-Tenant SFL System Scales on SFL Convergence Bound and Local Devices' Utility Gains.

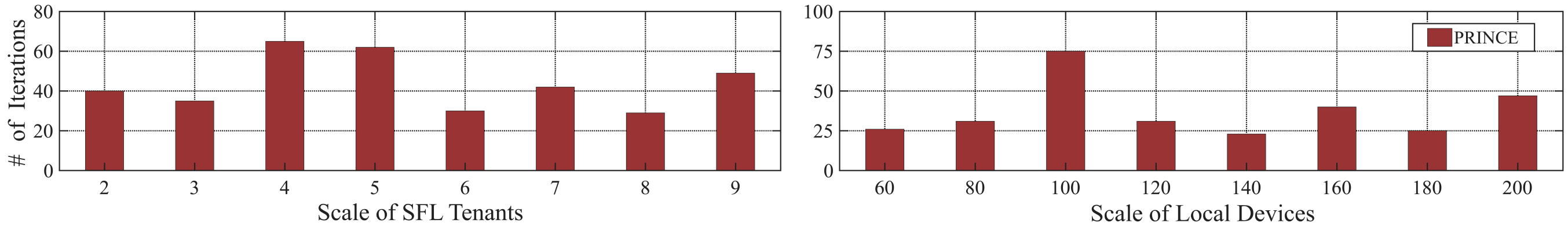

Fig. 6: Algorithmic Efficiency of our PRINCE mechanism under Different Multi-Tenant SFL System Scales.

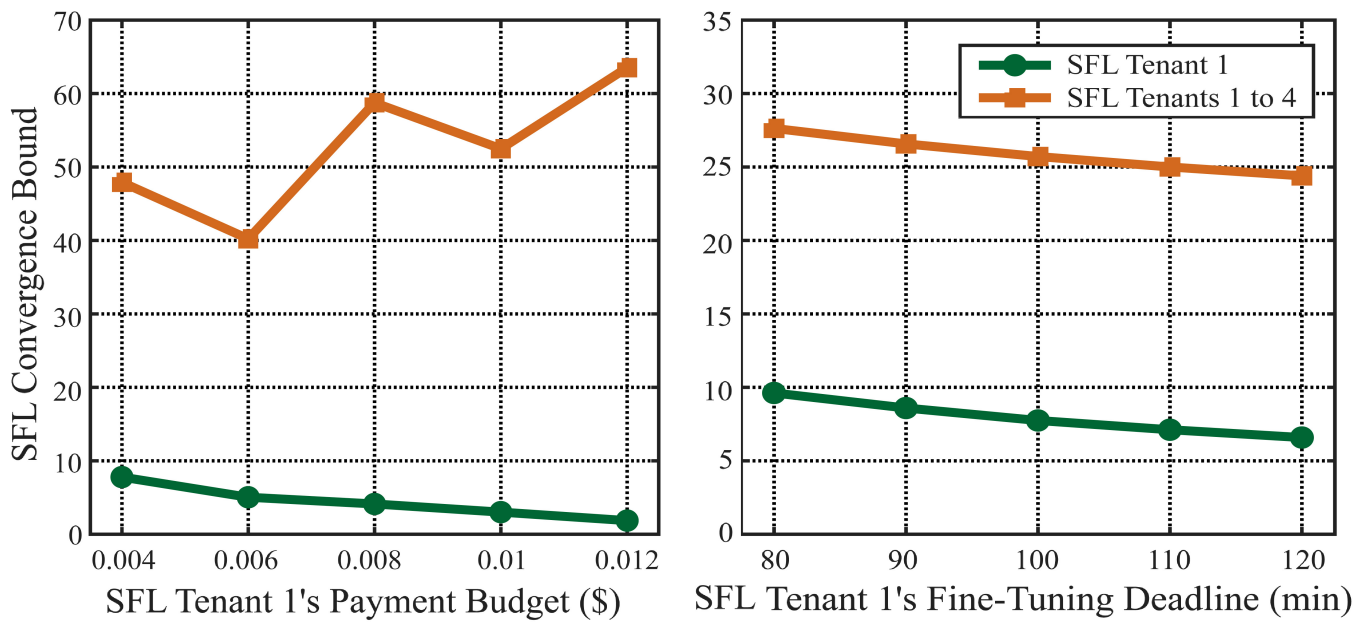

Fig. 7: Impact of Payment Budget and Fine-Tuning Deadline on PRINCE Performance.

## VII. Discussion

The proposed PRINCE mechanism could be extended to dynamic network edge environments, where local devices may join or leave the system, and the SFL tenant's FM fine-tuning workloads and payment budgets evolve over time. Specifically, PRINCE can operate in a time-slotted manner, allowing SFL tenants to adjust their incentive decisions on a rolling basis in response to dynamic system changes. When a local device joins or leaves, the system registrar is instantly updated, supporting SFL tenants in promptly adjusting their incentive-based client sampling strategies in the subsequent decision timeslots for efficient FM fine-tuning. Meanwhile, each SFL tenant regularly assesses the latest FM fine-tuning performance of its downstream task on a time-slotted basis, including the residual workload (i.e., the optimality gap) required for model convergence by the fine-tuning deadline. Accordingly, the SFL tenant adjusts its incentive strategy for device participation to ensure efficient split federated FM fine-tuning. Moreover, suppose an SFL tenant's payment budget changes over time, the SFL tenant correspondingly revises its incentive strategy for device participation to reflect the latest budget constraint.

## VIII. Conclusion

This paper investigates incentivizing multi-tenant SFL for FM fine-tuning at the network edge. The proposed incentive mechanism effectively coordinate self-interested local devices to participate in various SFL tenants' downstream tasks, thereby satisfying each SFL tenant's distinct FM fine-tuning requirements (e.g., FM types, performance targets, and fine-tuning deadlines). Specifically, we put forward a novel Price-Incentive Mechanism (PRINCE) for multi-tenant SFL, guiding multiple SFL tenants to offer strategic price incentives that solicit high-quality device participation, thereby optimizing their respective FM fine-tuning performance. We address the unique challenges of designing incentive mechanism in multi-tenant SFL environments, including independent device participation, device contribution assessment, and inter-tenant device competition. The superiority of PRINCE over state-of-the-art approaches is demonstrated through extensive simulation experiments featuring realistic FM fine-tuning workloads. This research lays a solid foundation for the widespread adoption of SFL in edge networks for FM fine-tuning.

# *Supplementary Material for* “Incentivizing Multi-Tenant Split Federated Learning for Foundation Models at the Network Edge”

Songyuan Li, Jia Hu, Geyong Min, and Haojun Huang

## Appendix A
## Proof of Theorem 1

*Proof.* Substituting Eq. (18) into Eq. (19), we obtain:

$$\begin{aligned}
&\mathbb{E}_{\mathcal{U}_i^{[k]}}[\mathbf{w}_i^{kI}\left(\mathbf{q}_i\right)] \\
=\ & \mathbf{w}_i^{(k-1)I} + \mathbb{E}\left[\sum\nolimits_{u_j\in\mathcal{U}_i^{[k]}} \frac{a_{i,j}}{q_{i,j}}\cdot\left(\mathbf{w}_{i,j}^{kI}-\mathbf{w}_i^{(k-1)I}\right)\right] \\
=\ & \mathbf{w}_i^{(k-1)I} + \sum_{j=1}^{N} q_{i,j}\cdot\frac{a_{i,j}}{q_{i,j}}\cdot\left(\mathbf{w}_{i,j}^{kI}-\mathbf{w}_i^{(k-1)I}\right) \\
=\ & \mathbf{w}_i^{(k-1)I} + \sum_{j=1}^{N} a_{i,j}\cdot\left(\mathbf{w}_{i,j}^{kI}-\mathbf{w}_i^{(k-1)I}\right) \\
=\ & \mathbf{w}_i^{(k-1)I} + \overline{\mathbf{w}}_i^{kI} - \mathbf{w}_i^{(k-1)I} \\
=\ & \overline{\mathbf{w}}_i^{kI},
\end{aligned} \tag{30}$$

where the fourth equation follows from the fact that $\sum_{u_j\in\mathcal{U}_i} a_{i,j}=1$. Therefore, we complete the proof by showing that $\mathbb{E}_{\mathcal{U}_i^{[k]}}[\mathbf{w}_i^{kI}]=\overline{\mathbf{w}}_i^{kI}$. This indicates that our proposed global SFL model aggregation scheme is bias-resilient with respect to the weighted aggregation model $\overline{\mathbf{w}}_i^{kI}=\sum_{j=1}^{N} a_{i,j}\cdot\mathbf{w}_{i,j}^{kI}$ with full device participation $\mathcal{U}$. □

## Appendix B
## Proof of Theorem 2

*Proof.* Following a similar argument of convergence analysis in [1], we first show that for any arbitrary device participation levels (probabilities) $q_{i,j}$, the variance between our obtained global SFL model $\mathbf{w}_i^{kI}$ in Eq. (18) and the global SFL model $\overline{\mathbf{w}}_i^{kI}$ under full device participation is bounded as:

$$\mathbb{E}\left\|\mathbf{w}_i^{kI}-\overline{\mathbf{w}}_i^{kI}\right\|^2 \le 4\cdot\gamma_{j,k}^2\cdot E^2\cdot\sum_{j=1}^{N}\frac{\left(1-q_{i,j}\right)a_{i,j}^2 G_{i,j}^2}{q_{i,j}}, \tag{31}$$

where $E$ is the number of stochastic gradient descent steps that a local device fine-tune on each data sample. Note that the key distinction of Eq. (31) from [1] is that our device participation levels $q_{i,j}$ are mutually independent, which does not rely on the assumption of $\sum_{j=1}^{N} q_{i,j}=1$ made in [1]. Note that, when $q_{i,j}=1$ for all local devices $u_j$, the variance in Eq. (31) is tightly bounded by zero. This is because, in such cases, our global SFL model $\mathbf{w}_i^{kI}$ in the left-hand side of Eq. (31) corresponds directly to the global SFL model $\overline{\mathbf{w}}_i^{kI}$ of full device participation. Here, $q_{i,j}=1$ implies full device participation for the SFL tenant $\psi_i$'s downstream task.

Meanwhile, we derive the SFL convergence bound under full device participation as follows:

$$\mathbb{E}\left[F\left(\overline{\mathbf{w}}_i^{kI}\right)\right]-F\left(\overline{\mathbf{w}}_i^{*}\right)\le\beta_i/k, \tag{32}$$

where $\beta_i$ is the same as defined in Eq. (21). Using mathematical induction, we also obtain a non-recursive bound on $\mathbb{E}\left\|\mathbf{w}_i^{kI}-\overline{\mathbf{w}}_i^{*}\right\|^2$, and demonstrate that its difference from the bound of full device participation $\mathbb{E}\left\|\overline{\mathbf{w}}_i^{kI}-\overline{\mathbf{w}}_i^{*}\right\|^2$, corresponds to the SFL variance introduced in Eq. (31).

Subsequently, we convert the bound on $\mathbb{E}_{\mathcal{U}_i^{[k]}}\left\|\mathbf{w}_i^{kI}-\overline{\mathbf{w}}_i^{*}\right\|^2$ into $\mathbb{E}\left[F\left(\mathbf{w}_i^{kI}\right)\right]-F\left(\overline{\mathbf{w}}_i^{*}\right)$ by leveraging the $L_i$-smoothness and $\mu_i$-Polyak-Lojasiewicz inequality properties of loss functions $F(\cdot)$. This yields an additional term of $\alpha_i\sum_{j=1}^{N}\left(1-q_{i,j}\right)a_{i,j}^2G_{i,j}^2/q_{i,j}$ in Eq. (21) compared to the SFL convergence bound Eq. (32) for full device participation. Specifically,

$$\mathbb{E}\left[F\left(\mathbf{w}_i^{kI}\right)\right]-F\left(\overline{\mathbf{w}}_i^{*}\right)\le\frac{1}{k}\left(\alpha_i\sum_{j=1}^{N}\frac{\left(1-q_{i,j}\right)a_{i,j}^2G_{i,j}^2}{q_{i,j}}+\beta_i\right) \tag{33}$$

Finally, substituting $k=K_i^{\Gamma}$ into Eq. (33), we arrive at Eq. (21). This concludes the proof. □

## Appendix C
## Proof of Theorem 3

*Proof.* Suppose an SFL tenant $\psi_i$ improves its pricing strategy from $\mathbf{P}_i$ to $\mathbf{P}_i'$, satisfying $\Lambda_i'(\mathbf{P}_i',\mathbf{P}_{\text{-}i}) < \Lambda_i'(\mathbf{P}_i,\mathbf{P}_{\text{-}i})$. To prove Theorem 3, we need to establish that $\Upsilon(\mathbf{P}_i',\mathbf{P}_{\text{-}i}) < \Upsilon(\mathbf{P}_i,\mathbf{P}_{\text{-}i})$.

The SFL tenant $\psi_i$ increases its pricing strategy to $\mathbf{P}_i'$ in order to incentivize higher device participation levels $\mathbf{q}_i$, aiming to reduce its expected global FM loss, as indicated by $\Lambda_i=\mathbb{E}\left[F\left(\mathbf{w}_i^{\Gamma}\left(\mathbf{q}_i\right)\right)\right]$. However, in the context of inter-SFL-tenant device competition, local devices that also participate in the downstream tasks of competing SFL tenants $\Psi\backslash\{\psi_i\}$, may reduce their participation levels in those downstream tasks due to the total SFL participation level constraint in Eq. (15a). This reduction in device participation levels could potentially

Songyuan Li, Jia Hu, and Geyong Min are with the Department of Computer Science, Faculty of Environment, Science and Economy, University of Exeter, Exeter EX4 4PY, U.K. (e-mail: S.Y.Li@exeter.ac.uk, J.Hu@exeter.ac.uk, G.Min@exeter.ac.uk).

Haojun Huang is with the School of Electronic Information and Communications, Huazhong University of Science and Technology, Wuhan 430074, China (e-mail: hjhuang@hust.edu.cn).

harm the FM fine-tuning performance of the competing SFL tenants, as represented by:

$$\sum_{\psi_v \in \Psi \setminus \{\psi_i\}} \Lambda'_v(\mathbf{P}'_i, \mathbf{P}_{\text{-}i}) > \sum_{\psi_v \in \Psi \setminus \{\psi_i\}} \Lambda'_v(\mathbf{P}_i, \mathbf{P}_{\text{-}i}). \tag{34}$$

According to our PRINCE mechanism design, only the pricing strategy improvements that result in an overall improvement in $\sum_{v=1}^{M} \Lambda'_v$ are approved. Mathematically, this condition is expressed as:

$$\Lambda'_i(\mathbf{P}'_i, \mathbf{P}_{\text{-}i}) - \Lambda'_i(\mathbf{P}_i, \mathbf{P}_{\text{-}i}) < \sum_{\psi_v \in \Psi \setminus \{\psi_i\}} [\Lambda'_v(\mathbf{P}_i, \mathbf{P}_{\text{-}i}) - \Lambda'_v(\mathbf{P}'_i, \mathbf{P}_{\text{-}i})]. \tag{35}$$

From this, it follows that:

$$\begin{aligned} &\Upsilon(\mathbf{P}'_i, \mathbf{P}_{\text{-}i}) - \Upsilon(\mathbf{P}_i, \mathbf{P}_{\text{-}i}) \\ &= [\Lambda'_i(\mathbf{P}'_i, \mathbf{P}_{\text{-}i}) + \sum_{\psi_v \in \Psi \setminus \{\psi_i\}} \Lambda'_v(\mathbf{P}'_i, \mathbf{P}_{\text{-}i})] - [\Lambda'_i(\mathbf{P}_i, \mathbf{P}_{\text{-}i}) + \sum_{\psi_v \in \Psi \setminus \{\psi_i\}} \Lambda'_v(\mathbf{P}_i, \mathbf{P}_{\text{-}i})] \\ &= [\Lambda'_i(\mathbf{P}'_i, \mathbf{P}_{\text{-}i}) - \Lambda'_i(\mathbf{P}_i, \mathbf{P}_{\text{-}i})] + \sum_{\psi_v \in \Psi \setminus \{\psi_i\}} [\Lambda'_v(\mathbf{P}'_i, \mathbf{P}_{\text{-}i}) - \Lambda'_v(\mathbf{P}_i, \mathbf{P}_{\text{-}i})] \\ &< 0. \end{aligned} \tag{36}$$

To summarize, $\Upsilon(\mathbf{P}'_i, \mathbf{P}_{\text{-}i}) < \Upsilon(\mathbf{P}_i, \mathbf{P}_{\text{-}i})$ always holds if $\Lambda'_i(\mathbf{P}'_i, \mathbf{P}_{\text{-}i}) < \Lambda'_i(\mathbf{P}_i, \mathbf{P}_{\text{-}i})$, confirming that our multi-SFL-tenant congestion game $\Omega$ is a potential game with $\Upsilon(\cdot)$ as its potential function. □

## Appendix D
## Proof of Theorem 4

*Proof.* Let $\{\mathbf{P}_i\}_{i=1}^{M}$ denote the pricing-strategy profile solved by the decentralized PRINCE mechanism, and $\{\mathbf{P}^*_i\}_{i=1}^{M}$ represent the optimal pricing-strategy profile that minimizes the total SFL convergence bound $\sum_{i=1}^{M} \Lambda'_i$.

We prove the Theorem 4 by reduction to absurdity. Assume that the pricing-strategy profile $\{\mathbf{P}_i\}_{i=1}^{M}$ does not minimize the total SFL convergence bound $\sum_{i=1}^{M} \Lambda'_i$, implying:

$$\sum_{i=1}^{M} \Lambda'_i(\mathbf{P}_i, \mathbf{P}_{\text{-}i}) > \sum_{i=1}^{M} \Lambda'_i(\mathbf{P}^*_i, \mathbf{P}^*_{\text{-}i}). \tag{37}$$

which suggests that there must exist a group of SFL tenants who achieve a smaller SFL convergence bound $\Lambda'_i$ under $\{\mathbf{P}^*_i\}_{i=1}^{M}$ than under $\{\mathbf{P}_i\}_{i=1}^{M}$. Consequently, the set of SFL tenants can be divided into groups, $\Psi_1$ and $\Psi_2$:

- **Group $\mathbf{\Psi_1}$**: Comprises the SFL tenants that experience a smaller SFL convergence bound under $\{\mathbf{P}^*_i\}_{i=1}^{M}$ than under $\{\mathbf{P}_i\}_{i=1}^{M}$, implying:

$$\sum_{\psi_i \in \Psi_1} \Lambda'_i(\mathbf{P}_i, \mathbf{P}_{\text{-}i}) > \sum_{\psi_i \in \Psi_1} \Lambda'_i(\mathbf{P}^*_i, \mathbf{P}^*_{\text{-}i}). \tag{38}$$

The decrement in the SFL convergence bound for these SFL tenants in $\Psi_1$ is formulated as:

$$\Delta D = \sum_{\psi_i \in \Psi_1} (\Lambda'_i(\mathbf{P}_i, \mathbf{P}_{\text{-}i}) - \Lambda'_i(\mathbf{P}^*_i, \mathbf{P}^*_{\text{-}i})). \tag{39}$$

- **Group $\mathbf{\Psi_2}$**: Comprises the remaining SFL tenants, for whom the SFL convergence bound either remains unchanged or increases. In the worst case, their SFL convergence bound increases from $\{\mathbf{P}_i\}_{i=1}^{M}$ to $\{\mathbf{P}^*_i\}_{i=1}^{M}$, implying:

$$\sum_{\psi_i \in \Psi_2} \Lambda'_i(\mathbf{P}_i, \mathbf{P}_{\text{-}i}) \leq \sum_{\psi_i \in \Psi_2} \Lambda'_i(\mathbf{P}^*_i, \mathbf{P}^*_{\text{-}i}). \tag{40}$$

The increment in the SFL convergence bound for these SFL tenants in $\Psi_2$ is formulated as:

$$\Delta I = \sum_{\psi_i \in \Psi_2} (\Lambda'_i(\mathbf{P}^*_i, \mathbf{P}^*_{\text{-}i}) - \Lambda'_i(\mathbf{P}_i, \mathbf{P}_{\text{-}i})). \tag{41}$$

The assumption in Eq. (37) indicates $\Delta D > \Delta I$, triggering the pricing-strategy adjustment condition (see *Line 12* of Algorithm 2) in our PRINCE algorithm, further minimizing the total SFL convergence bound. It is important to note that the PRINCE algorithm does not terminate its strategy-improvement iterations until no SFL tenant can update its pricing strategy to acheive an overall improvement in $\sum_{v=1}^{M} \Lambda'_v$. In other words, $\{\mathbf{P}_i\}_{i=1}^{M}$ cannot be the finalized pricing-strategy profile obtained by the PRINCE mechanism, which contradicts the initial assumption. This concludes the proof. □